\documentclass[11pt]{article}

\PassOptionsToPackage{table}{xcolor}
\usepackage[final]{acl}

\usepackage{times}
\usepackage{latexsym}
\usepackage{enumitem}
\usepackage{multirow}
\usepackage{makecell}
\usepackage{microtype}
\usepackage{titletoc}
\usepackage[T1]{fontenc}
\usepackage[utf8]{inputenc}
\usepackage{inconsolata}
\usepackage{graphicx}
\usepackage{booktabs}
\usepackage{tipa}
\usepackage{pifont}
\usepackage{subcaption} 
\usepackage{amsmath} 
\usepackage{amssymb} 
\usepackage{colortbl}
\usepackage{pifont}
\newcommand{\cmark}{\textcolor{green!60!black}{\ding{51}}}
\newcommand{\xmark}{\textcolor{red}{\ding{55}}}
\usepackage{xcolor}
\definecolor{skybluecolor}{RGB}{173, 216, 230}

\usepackage[breakable,skins]{tcolorbox}

\newcommand{\gup}[1]{\textcolor{green!60!black}{$\uparrow$\,}#1}
\newcommand{\gdn}[1]{\textcolor{red}{$\downarrow$\,}#1}

\title{SLATE: Are AI-Generated Slides Educationally Effective? A Benchmark for Language Teaching Quality and Learner Knowledge Acquisition}

\author{
\textbf{Jingzhuo Wu\textsuperscript{1}}\quad
\textbf{Jiajun Zhang\textsuperscript{2,3}}\quad
\textbf{Yi Liu\textsuperscript{4}}\quad
\textbf{Leqi Zheng\textsuperscript{5}}
\\
\textbf{Yuheng Jing\textsuperscript{3}}\quad
\textbf{Xinyuan Zhou\textsuperscript{1}}\quad
\textbf{Quan Yang\textsuperscript{1}\thanks{Corresponding author.}}
\\[0.4em]
\textsuperscript{1}Beijing Normal University
\quad
\textsuperscript{2}University of Science and Technology of China
\\
\textsuperscript{3}Institute of Automation, Chinese Academy of Sciences
\\
\textsuperscript{4}Beijing Language and Culture University
\quad
\textsuperscript{5}Tsinghua University
\\[0.3em]
\texttt{jingzhuowu923@gmail.com}
}

\begin{document}
\maketitle

\begin{abstract}
LLMs have achieved remarkable capabilities in generating language teaching slides. However, a critical mismatch persists between visual polish and actual instructional effectiveness. To address this gap, we introduce \textbf{\texttt{SLATE}} (\textbf{S}lide-based \textbf{L}earning \textbf{A}ssessment for \textbf{T}eaching \textbf{E}ffectiveness), the first benchmark that evaluates AI-generated language teaching slides through instructional effectiveness and learner knowledge acquisition. \textbf{\texttt{SLATE}} transforms linguistics olympiad puzzles from low-resource languages with negligible web presence into 90 standardized instructional units comprising 1,133 assessable items, paired with a structured course outline and matched near- and far-transfer test sets. This pretest--posttest design eliminates pretrained knowledge leakage, ensuring gains reflect learning rather than prior recall. Using VLMs as scalable learner proxies and directionally supported by a three-system human pilot, our results show that content validity exhibits a weak association with learning gain, while pedagogical design exhibits a robust positive association. Moreover, most systems show a significant gap between near- and far-transfer accuracy, and even frontier models can produce negative learning gains. \textbf{\texttt{SLATE}} reveals a dissociation between artifact quality and instructional effectiveness, calling for a paradigm shift in how generative teaching systems are built, evaluated, and deployed.
\end{abstract}


\section{Introduction}
Large language models have revolutionized AI slide generation. Frontier systems such as
PPTAgent~\citep{PPTAgent},
AutoPresent~\citep{AutoPresent}, and
SlideBot~\citep{SlideBot} produce visually polished, professionally structured teaching slides. AI-generated slides have
become a core application in intelligent education~\citep{SlideGen}, and are widely adopted by language
teachers for lesson preparation and classroom instruction.
Corresponding evaluation benchmarks such as PPTEval~\citep{PPTEval}, SlidesGen-Bench~\citep{SlidesGen-Bench}, and PresentBench~\citep{PresentBench} assess artifact-level properties including content coherence, visual design, and structural completeness~\citep{Unifying}.

\begin{figure*}[t]
\centering
\includegraphics[width=\textwidth]{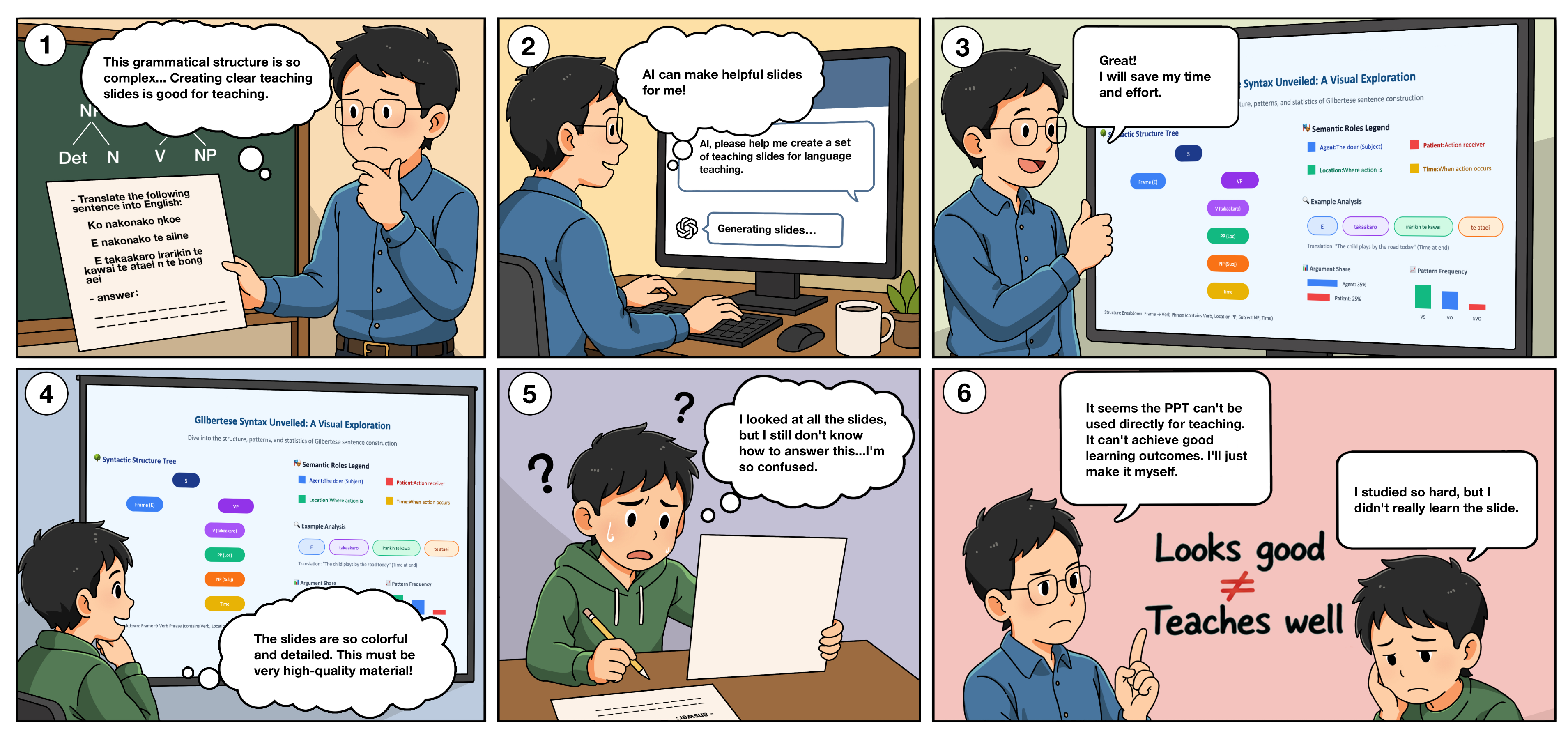}
\vspace{-1.5em}
\caption{AI-generated slides can appear visually polished and
pedagogically plausible, yet still fail to help learners
acquire target concepts or solve novel problems after study.}
\vspace{-0.5em}
\label{fig:comic}
\end{figure*}

\begin{table}[t]
\centering
\setlength{\tabcolsep}{3pt}
\resizebox{\linewidth}{!}{
\begin{tabular}{l cc cccc}
\toprule
& \multicolumn{2}{c}{\textbf{Scope}} 
& \multicolumn{4}{c}{\textbf{Evaluation}} \\
\cmidrule(lr){2-3}\cmidrule(lr){4-7}
\textbf{Benchmark} 
& \textbf{Slide} & \textbf{Lang.}
& \textbf{Artifact} & \textbf{Pedagog.} & \textbf{Learner} & \textbf{Transfer} \\
& \textbf{gen.} & \textbf{teaching}
& \textbf{quality} & \textbf{grounding} & \textbf{outcomes} & \textbf{eval.} \\
\midrule
PPTEval~\citep{PPTEval}
  & \cmark & \xmark & \cmark & \xmark & \xmark & \xmark \\
SlidesGen-Bench~\citep{SlidesGen-Bench}
  & \cmark & \xmark & \cmark & \xmark & \xmark & \xmark \\
PresentBench~\citep{PresentBench}
  & \cmark & \xmark & \cmark & \xmark & \xmark & \xmark \\
TutorBench~\citep{TutorBench}
  & \xmark & \cmark & \xmark & \cmark & \xmark & \xmark \\
MathTutorBench~\citep{MathTutorBench}
  & \xmark & \cmark & \xmark & \cmark & \xmark & \xmark \\
\rowcolor{gray!15}
\textbf{SLATE (Ours)}
  & \cmark & \cmark & \cmark & \cmark & \cmark & \cmark \\
\bottomrule
\end{tabular}}
\vspace{-0.5em}
\caption{Comparison of \textbf{\texttt{SLATE}} with existing slide generation and educational benchmarks.}
\label{tab:benchmark-comparison}
\vspace{-1em}
\end{table}

However, many AI-generated language-teaching slides that
perform well on these artifact metrics show highly variable instructional outcomes in practice~\citep{PAIGE}. Decks
generated directly from frontier systems often fail to match
learners' cognitive pace~\citep{Beyond} or align with established patterns of
second language acquisition~\citep{MathVC}. More critically, existing evaluation
paradigms suffer from a fundamental limitation: they evaluate
slides solely on surface-level properties,
without ever measuring learner knowledge acquisition,
conceptual understanding, or knowledge transfer~\citep{EducationQ}. This creates
a paradox illustrated in Figure~\ref{fig:comic}: a visually polished and structurally
complete teaching deck may produce no effective learning
gain, or even result in negative learning outcomes.

To close this gap, we introduce \textbf{\texttt{SLATE}}
(\textbf{S}lide-based \textbf{L}earning \textbf{A}ssessment
for \textbf{T}eaching \textbf{E}ffectiveness), the first
benchmark dedicated to evaluating AI-generated
language-teaching slides through instructional effectiveness
and learner transfer. As shown in
Table~\ref{tab:benchmark-comparison}, \textbf{\texttt{SLATE}}
is the first to unify learning gain measurement, near/far-transfer
evaluation, pretraining leakage control, and human validation
within a single assessment framework. Grounded in cognitive load theory,
multimedia learning theory, and transfer taxonomy,
\textbf{\texttt{SLATE}} evaluates decks along four
complementary, theory-driven dimensions: linguistic knowledge validity (D1),
cognitive-pedagogical design (D2), visual communication
quality (D3), and learning and transfer outcomes (D4).
\textbf{\texttt{SLATE}} transforms linguistics olympiad puzzles
from low-resource languages ~\citep{LINGOLY} into 90 standardized instructional units comprising 1,133
assessable items, and structured course outlines and matched near- and far-transfer test sets.

Evaluating 10 frontier systems on \textbf{\texttt{SLATE}}
reveals a systematic dissociation between artifact quality and
instructional effectiveness: content validity correlates weakly
with learning gain, while pedagogical design correlates strongly. All systems show a near/far transfer
gap, supporting local pattern reuse but failing to scaffold generalization. Critically, several systems produce
\emph{negative} learning gains, a failure mode largely
undetectable by artifact-based rubrics. These findings are supported by a 30-participant three-system human pilot (\S\ref{sec:VLM_Student_Validity}) and a reliability analysis
(\S\ref{sec:reliability}) showing high agreement between our
multi-agent framework and expert judgments. Quantitative
and qualitative analyses (\S\ref{sec:analysis}) unpack these
patterns and mechanisms.

Our core contributions are threefold:
\ding{182} First, we introduce \textbf{\texttt{SLATE}}, the first benchmark to evaluate AI-generated language-teaching slides via instructional effectiveness and learner transfer, featuring a theory-grounded four-dimensional framework (D1--D4) and 90 standardized instructional units.
\ding{183} We establish a controlled evaluation protocol
grounded in low-resource language tasks, mitigating pretraining
knowledge leakage via a pretest--study--posttest--transfer
pipeline validated against human learners with 85.7\% directional agreement and perfect cross-system ranking consistency ($n{=}30$).
\ding{184} We empirically reveal a systematic dissociation
between artifact-level quality and learning outcomes
across 10 frontier systems: pedagogical design (D2) rather
than content accuracy (D1) dominates instructional
effectiveness, with several systems producing negative gains
invisible to existing rubrics.

\section{Related Work}

\paragraph{Slide Generation and Evaluation.}
Driven by rapid advances in multimodal generative AI~\citep{realchart2code},
slide generation has advanced from template-based summarization~\citep{DOC2PPT}
to sophisticated agentic pipelines.
PPTAgent~\citep{PPTAgent} infers functional schemas from reference decks;
AutoPresent~\citep{AutoPresent} generates Python-pptx code directly;
SlidesGen~\citep{SlideGen} decomposes creation into coordinated subtasks; SlideBot~\citep{SlideBot} incorporates principles from learning science
into slide planning; and SlideTailor~\citep{slidetailor} personalizes
scientific-deck generation.
Existing evaluation frameworks---PPTEval~\citep{PPTEval},
SlidesGen-Bench~\citep{SlidesGen-Bench}, and PresentBench~\citep{PresentBench}---%
score content, design, and coherence as artifact properties ~\citep{PAIGE}.
All existing work treats slides as \emph{artifacts}:
none evaluates whether studying a generated deck improves learner outcomes.

\paragraph{Educational Benchmarks and Learning Transfer.}
Most educational benchmarks position LLMs as \emph{students}:
LINGOLY~\citep{LINGOLY}, ModeLing~\citep{modeLing},
PuzzLing~\citep{PuzzLingMachines}, AGIEval~\citep{AGIEval},
and MMLU~\citep{Measuring} test reasoning and knowledge mastery
but do not evaluate whether models can \emph{generate} effective teaching materials.
TutorBench~\citep{TutorBench} benchmarks interactive tutoring quality,
yet still targets model behavior rather than artifact effectiveness ~\citep{Agent4Edu}.
\textbf{\texttt{SLATE}} shifts this paradigm by evaluating generated artifacts
through a pretest--study--posttest--transfer protocol grounded in
transfer taxonomy~\citep{Barnett},
which predicts that far transfer requires explicit abstraction support
beyond local pattern reuse~\citep{Transfer}. A comprehensive review of related work is provided in Appendix~\ref{app:Additional_Related_Work}.

\section{The \textbf{\texttt{SLATE}} Benchmark}

\subsection{Task Definition}

\textbf{\texttt{SLATE}} formulates language-teaching slide generation as a conditional instructional artifact generation task. Given a puzzle-derived instructional unit $U$ and a structured course outline $O$, a generation model $\mathcal{G}$ produces a slide deck $S = \mathcal{G}(U, O)$ whose goal is to teach the target linguistic concept. A successful deck must preserve linguistic correctness, organize the concept into a learnable sequence, provide visual support for rule induction, and improve learner performance on source and transfer items.

This formulation differs from standard text-to-slide generation in two ways. First, the source is an instructional unit distilled from a rule-learning problem, not a document. Second, the output is evaluated through learner-side outcomes in addition to artifact quality.

\subsection{Benchmark Overview}

\begin{figure*}[t]
\centering
\includegraphics[width=0.96\textwidth]{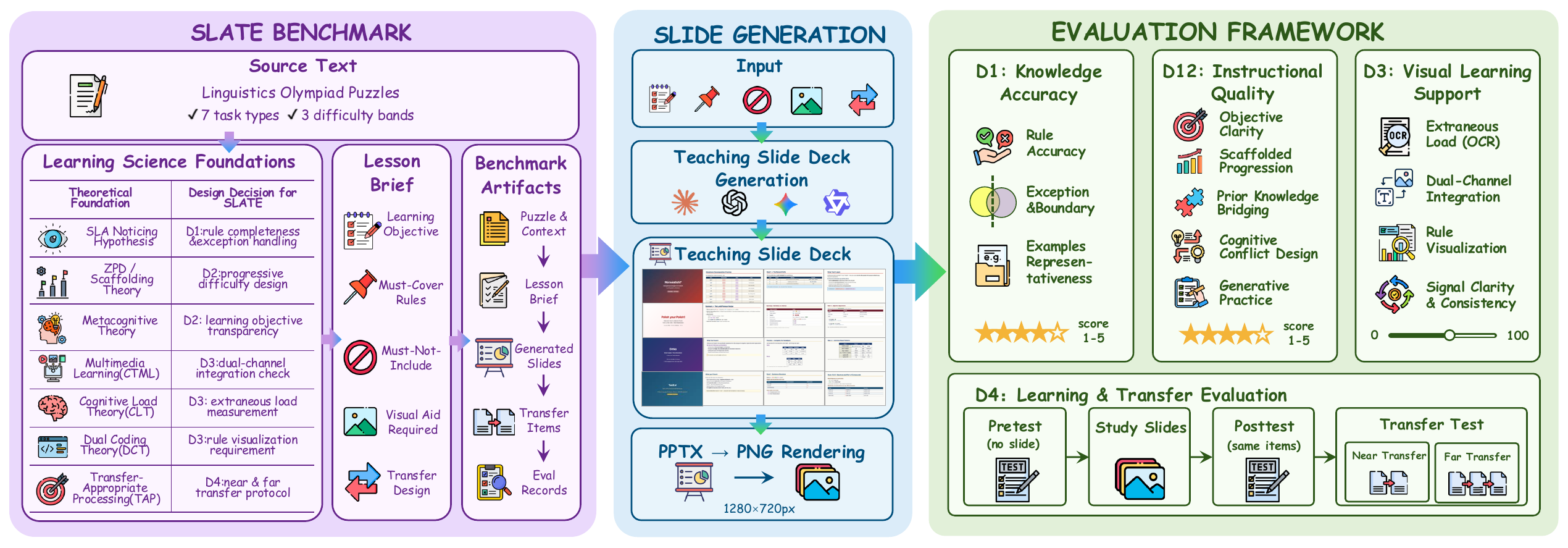}
\vspace{-1em}
\caption{Overview of \textbf{\texttt{SLATE}}. Source linguistics olympiad puzzles are transformed into instructional units; generation models produce teaching decks; D1--D4 evaluate content, pedagogy, visual support, and learner transfer.}
\vspace{-0.5em}
\label{fig:main_results_overview}
\end{figure*}

\begin{table}[t]
\centering
\small
\begin{tabular}{lr}
\toprule
\textbf{Property} & \textbf{Value} \\
\midrule
Instructional units & 90 \\
Assessment sub-items & 1{,}133 \\
Task types & 7 \\
Difficulty bands & 3 \\
\quad Basic / Intermediate / Challenge & 2 / 70 / 18 \\
Avg.\ teachable rules per unit & 5.8 \\
Near-transfer items per unit & 2 \\
Far-transfer items per unit & 2 \\
Generation systems evaluated & 10 \\
Evaluation dimensions & 4 (D1--D4) \\
\bottomrule
\end{tabular}
\caption{SLATE benchmark statistics.}
\label{tab:benchmark_stats}
\vspace{-1.2em}
\end{table}

\textbf{\texttt{SLATE}} covers 7 linguistic task types: paradigm completion (37), translation (31), morphophonology (10), numeral systems (6), syntax/word order (4), error detection (1), and language identification (1). We select linguistics olympiad puzzles for three methodological reasons that jointly address key validity threats in educational AI evaluation. First, all puzzles are drawn from low-resource, endangered, or constructed languages with negligible internet presence, ensuring that the macro pretest baseline (23.78\%) reflects genuine zero-knowledge performance rather than prior exposure; any post-study gain can therefore be attributed to the slide deck rather than recalled knowledge. Second, the puzzle-answer pairs are rarely indexed in public corpora, substantially reducing the risk that VLM student models retrieve solutions from pretraining memory rather than learning from instruction. Third, the rule-induction structure of each puzzle---inferring generalizations from a finite example set and applying them to novel inputs---natively instantiates the near- and far-transfer distinction central to our evaluation protocol~\citep{Barnett}, allowing transfer distance to be operationalized precisely rather than approximated. These categories span diverse instructional demands, from local contrastive explanation to compositional abstraction. Each unit is scored on five factors (rule complexity, exception density, representation density, transfer abstraction, distractor strength) and assigned to one of three difficulty bands. Table~\ref{tab:benchmark_stats} summarizes the benchmark statistics.

\subsection{Construction Pipeline}

The benchmark is constructed through a four-stage pipeline, illustrated in Figure~\ref{fig:main_results_overview}.

\paragraph{Stage 1: Source Puzzle Parsing.}
Raw linguistics olympiad puzzles are parsed into structured representations containing language name, task type, context, sub-items, and answers. This produces 1{,}133 assessable sub-items across 90 puzzles.

\paragraph{Stage 2: Instructional-Unit Construction.}
Each puzzle is converted from a problem-solving format into a concept-centered teaching unit by trained annotators. Annotators extract 2--6 teachable linguistic rules per puzzle (morphological alternations, agreement patterns, structural templates, etc.), assess instructional difficulty on a five-dimensional rubric, and generate matched near-transfer and far-transfer items. Each unit stores the target rules, key terms, exceptions, representative examples, and transfer items. Annotator qualifications and inter-annotator agreement are reported in Appendix~\ref{app:quality_control}.

\paragraph{Stage 3: Course-Outline Generation.}
For each unit, annotators compile a structured outline as the generation specification. The outline contains five mandatory fields: \textbf{Learning Objective}, \textbf{Must-Cover Rules}, \textbf{Must-Not-Include} (prohibiting answer exposure), \textbf{Visual Aid Required}, and \textbf{Transfer Design}. Transfer answers are never exposed to the generator, preventing evaluation leakage.

\paragraph{Stage 4: Slide Generation.}
Each generation system receives the same course outline and produces a self-contained HTML teaching deck, which is rendered to PNG pages via Playwright for downstream evaluation. We standardize on HTML to ensure all systems produce output in a uniform, programmatically renderable format, enabling consistent visual evaluation across systems. Slide count targets are difficulty-dependent (Basic: 8--12, Intermediate: 10--14, Challenge: 12--16 pages).

\subsection{Evaluation Framework}

\textbf{\texttt{SLATE}} evaluates generated decks along four theory-grounded dimensions.

\paragraph{D1: Linguistic Knowledge Validity.}
D1 assesses whether a deck teaches the intended content correctly, covering rule coverage, data fidelity, metalinguistic accuracy, answer non-leakage, and scope appropriateness.

\paragraph{D2: Cognitive-Pedagogical Design.}
D2 evaluates instructional organization: cognitive load management, worked examples, scaffolded progression, dual-channel integration, and practice opportunities. Both D1 and D2 are grounded in Cognitive Load Theory~\citep{Cognitive} and the Cognitive Theory of Multimedia Learning~\citep{Mayer_2005}.

\paragraph{D3: Visual Communication Quality.}
D3 provides lightweight visual-support diagnostics computed automatically from rendered deck properties, measuring slide count, text density, table usage, color diversity, visual aid completeness, and exercise inclusion. Unlike D1 and D2, D3 does not assess perceptual or cognitive dimensions of visual design; it serves as a coarse filter for identifying decks that lack basic visual scaffolding.

\paragraph{D4: Learning and Transfer.}
D4 measures learner-side outcomes through a four-stage protocol. The student model first answers source items without instruction (pretest), studies the slide deck as PNG pages, answers the same items again (posttest), and then answers matched near-transfer and far-transfer items. Near-transfer items require applying taught rules to new inputs within the same structural pattern; far-transfer items require generalization across knowledge domain or functional context~\citep{Barnett}. Reported metrics include posttest gain, near-transfer accuracy, and far-transfer accuracy.

\section{Experiments}

\subsection{Experimental Setup}

\paragraph{Generation systems.}
We evaluate 10 frontier LLM systems under a unified setup: Claude Opus 4.7, Claude Sonnet 4.6, GPT-5.4, Gemini-3.1-Pro, Qwen3.6-Max, Qwen3.6-Plus, DeepSeek-V4-Pro, MiniMax-M2.7, Kimi-K2.6, and GLM-5.1. Each system receives the same course outline for each of the 90 instructional units and generates a self-contained HTML teaching deck. All systems complete all 90 units.

\paragraph{Student model and D4 protocol.}
D4 uses Qwen3-VL-8B-Instruct as the student model, selected based on the ablation study (Section\S~\ref{sec:ablation}), which evaluates four VL models of varying capacity (see Appendix~\ref{app:model_mapping} for full model names) and confirms that the 8B model balances sensitivity to instructional input with sufficient baseline capability. For each unit, the student answers source items without instruction (pretest), studies the rendered slide deck as PNG pages, answers the same items again (posttest), and finally answers near-transfer and far-transfer items. We report puzzle-level macro averages. The pretest baseline is 23.78\%. The cross-system ranking validity of this VLM proxy is validated via a three-system human pilot (§\ref{sec:VLM_Student_Validity}, Appendix~\ref{app:human_pilot}, Appendix~\ref{app:Human_Study}).

\paragraph{D1/D2 multi-agent evaluation.}
D1 and D2 are scored by a multi-agent panel of three heterogeneous LLMs (Claude Opus 4.7, Gemini-3.1-Pro, and GPT-5.4), each independently rating all 10 sub-dimensions (D1.1--D1.5, D2.1--D2.5) on a 1--5 scale. Final scores are averaged across the three judges to reduce single-model bias. Each reported score represents the mean of 5 independent runs per judge.

To validate this setup, three linguistics experts (all with doctoral training; two with low-resource language experience; one with $>$3 years of language-teaching and slide-design experience) independently rated a stratified sample of 29 decks after rubric calibration. We report inter-rater reliability and human--model agreement in Section\S~\ref{sec:discussion}.

\paragraph{D3 evaluation.}
D3 is computed automatically from rendered HTML properties: slide count, words per slide (WPS), table count, color diversity, visual aid completeness (VA\%), and exercise inclusion rate.

\subsection{Main Results}

Table~\ref{tab:main_results} presents the full results across all four dimensions. On D4, Claude Opus 4.7 and Qwen3.6-Max achieve the largest positive gains (\gup{5.46}\% and \gup{5.14}\%), followed by MiniMax-M2.7 (\gup{3.27}\%) and GPT-5.4 (\gup{2.61}\%). Four systems produce negative gains, confirming that SLATE discriminates systems on instructional effectiveness rather than artifact plausibility alone.

\begin{table*}[!t]
\small
\centering
\setlength{\tabcolsep}{5pt}
\begin{tabular}{l|cc|c|cccc}
\toprule
\multirow{2}{*}{\textbf{System}}
& \multicolumn{2}{c|}{\textbf{Artifact Quality}}
& \textbf{D3}
& \multicolumn{4}{c}{\textbf{D4: Learning \& Transfer}} \\
& D1 & D2 & VA\%
& Gain (\%) & Post & Near & Far \\
\midrule
Claude Opus 4.7 ~\citep{claudeOpus4.7} & \textbf{4.19} & \textbf{4.42} & \textbf{99}
& \textbf{\gup{5.46}} & \textbf{29.25} & \textbf{41.67} & 12.22 \\
Qwen3.6-Max ~\citep{qwen2026qwen36} & 3.33 & 3.95 & \textbf{99}
& \underline{\gup{5.14}} & \underline{28.92} & 39.89 & 12.36 \\
MiniMax-M2.7 ~\citep{minimax2026m27}& 3.77 & 3.91 & \textbf{99}
& \gup{3.27} & 27.06 & 38.33 & 15.00 \\
GPT-5.4  ~\citep{openai2026gpt54}& \underline{4.20} & \underline{4.24} & \textbf{99}
& \gup{2.61} & 26.40 & 35.00 & 13.33 \\
Claude Sonnet 4.6 ~\citep{claudeSonnet4.6}& 4.17 & 3.72 & 98
& \gup{1.85} & 25.63 & \textbf{41.67} & 12.22 \\
Qwen3.6-Plus ~\citep{qwen2026qwen36}& 4.01 & 4.23 & \textbf{99}
& \gup{0.38} & 24.17 & \underline{41.11} & 13.33 \\
Kimi-K2.6  ~\citep{kimik26}& 4.01 & 4.13 & 91
& \gdn{0.89} & 22.90 & 36.11 & \underline{15.56} \\
GLM-5.1  ~\citep{glm5team2026glm5vibecodingagentic}& 4.05 & 3.60 & 80
& \gdn{1.24} & 22.54 & 31.67 & 13.89 \\
DeepSeek-V4-Pro  ~\citep{deepseek2026v4}& 3.11 & 2.02 & 61
& \gdn{1.68} & 22.10 & 35.39 & 11.24 \\
Gemini-3.1-Pro  ~\citep{gemini31pro}& 3.60 & 2.88 & 79
& \gdn{2.51} & 21.28 & 32.02 & \textbf{16.29} \\
\bottomrule
\end{tabular}
\vspace{-0.5em}
\caption{\textbf{Main results on \textbf{\texttt{SLATE}}} (sorted by D4 gain). D1/D2 are multi-agent judge averages (1--5 scale; 5 runs $\times$ 3 judges). VA\%: visual aid completeness (full D3 breakdown in Appendix). D4 gain is measured against the pretest baseline of 23.78\%. \textbf{Bold}: best; \underline{underline}: second best.}
\label{tab:main_results}
\vspace{-1.2em}
\end{table*}

The most striking pattern in Table~\ref{tab:main_results} is the dissociation between artifact quality and learning outcomes. GPT-5.4 scores highest on D1 (4.20) yet ranks only fourth on gain; MiniMax-M2.7 scores below average on D1 (3.77) yet ranks third. This dissociation is not explained by visual design alone, though visual completeness appears necessary: all systems with VA\%$<$80\% produce negative gains. Across all systems, near-transfer scores (31--42\%) consistently exceed far-transfer scores (11--16\%), indicating that current decks support pattern reuse more effectively than conceptual generalization.

\subsection{Ablation Study}
\label{sec:ablation}

To isolate the contribution of slides from other factors, we conduct a controlled ablation with four student VLMs of increasing capacity across six conditions (Table~\ref{tab:ablation_results}).

\begin{table}[t]
\small
\centering
\setlength{\tabcolsep}{2pt}
\begin{tabular}{ll|cccc}
\toprule
& & \multicolumn{4}{c}{\textbf{Student Model}} \\
\textbf{Cond.} & \textbf{Setting} & Q25-3B & Q3-4B & Q25-7B & Q3-8B \\
\midrule
A & Ctx only & 14.6 & 19.9 & 16.9 & 22.9 \\
B & Ctx + slides & 13.1 & 24.2 & 20.9 & 28.0 \\
V1 & No input & 9.0 & 9.2 & 8.0 & 10.4 \\
V2 & Slides only & 13.1 & 17.7 & 16.1 & 20.4 \\
V3 & Ctx + 3-frame & 13.3 & 20.8 & 17.7 & 24.4 \\
V4 & 3-frame only & 12.0 & 12.6 & 12.1 & 13.8 \\
\midrule
\multicolumn{6}{l}{\textit{Effect decomposition (\%)}} \\
\midrule
B$-$A & Slide effect & \gdn{1.4} & \gup{4.3} & \gup{4.0} & \gup{5.0} \\
A$-$V1 & Context effect & \gup{5.6} & \gup{10.7} & \gup{8.8} & \gup{12.5} \\
V2$-$V1 & Slide compens. & \gup{4.1} & \gup{8.5} & \gup{8.1} & \gup{10.0} \\
B$-$V3 & Full vs.\ 3-fr. & \gdn{0.2} & \gup{3.4} & \gup{3.1} & \gup{3.5} \\
\bottomrule
\end{tabular}
\caption{\textbf{Ablation study} across 4 student VLMs and 6 conditions using Claude Opus 4.7 slides. Upper block: accuracy (\%); lower block: effect sizes (\%).}
\label{tab:ablation_results}
\end{table}

\begin{figure}[t]
\centering
\includegraphics[width=\columnwidth]{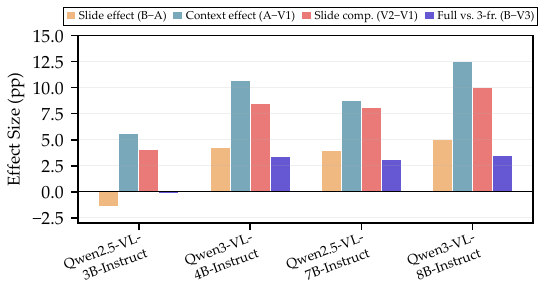}
\vspace{-1em}
\caption{Ablation effect decomposition across 4 student VLMs. Context is the strongest factor; slide effects scale with model capacity; full-frame consistently outperforms 3-frame sampling.}
\label{fig:ablation_effects}
\vspace{-1.5em}
\end{figure}

Three findings stand out. First, slides produce positive learning gain for 3 of 4 student models; only the smallest model (Qwen2.5-VL-3B) shows negative slide effect (\gdn{1.4}\%). This is consistent with the expertise reversal effect in Cognitive Load Theory~\citep{Expertise}: instructional scaffolding that benefits intermediate learners can impose extraneous cognitive load on learners whose processing capacity is insufficient to integrate multi-page visual input. The monotonic scaling of slide effects across the four models (\gdn{1.4}, \gup{4.3}, \gup{4.0}, \gup{5.0}\%) rules out the possibility that the 8B result is an isolated artifact: the trend is systematic and capacity-dependent. Second, puzzle context is the strongest single factor (\gup{5.6} to \gup{12.5}\%), but slides alone can partially compensate for missing context (\gup{4.1} to \gup{10.0}\% gain over zero-information baseline). Third, full-frame presentation outperforms 3-frame sampling by $\sim$3\%, confirming that intermediate slides contain valuable instructional content. Figure~\ref{fig:ablation_effects} visualizes these effect decompositions across four student models.

\subsection{Analysis}
\label{sec:analysis}

\begin{figure}[t]
\centering
\vspace{-0.5em}
\includegraphics[width=\columnwidth]{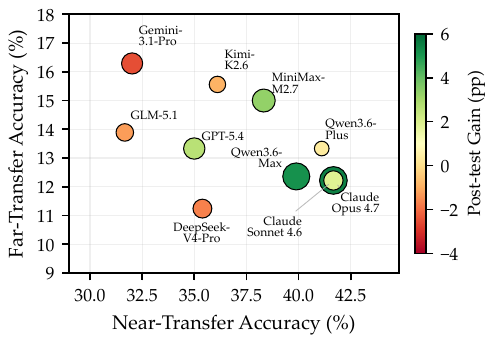}
\caption{Near- vs.\ far-transfer accuracy for each system (color = posttest gain, size = $|$gain$|$). Systems cluster in a high-near / low-far region, indicating a systematic gap between pattern reuse and conceptual generalization.}
\label{fig:transfer_gap}
\vspace{-0.5em}
\end{figure}

\paragraph{High near-transfer, weak far-transfer.}
Across systems, near-transfer accuracy ($\sim$35--42\%) substantially exceeds far-transfer ($\sim$11--16\%). This 20+\% gap persists even for the strongest systems, suggesting that current decks help learners recognize and reuse local patterns but do not reliably support rule abstraction for novel contexts. Figure~\ref{fig:transfer_gap} visualizes this pattern: systems cluster in the lower-right region, with strong near-transfer but uniformly weak far-transfer regardless of gain.

\paragraph{Task-type stratified gains.}
We group the 90 units into four macro categories: Paradigm Completion (37), Translation (31), Morphophonology (10), and Other (12; numeral systems, syntax, error detection, language ID). We report mean gain per category (Table~\ref{tab:tasktype_results}). Paradigm completion is the most teachable category (\gup{3.5}\% mean gain; 47.2\% near-transfer), likely because paradigm tasks present explicit input-output mappings that naturally lend themselves to tabular visualization and pattern-based instruction. Translation tasks show the weakest gains (\gdn{1.1}\%), consistent with their reliance on contextual inference and compositional semantics that are harder to scaffold within static visual materials. The near/far transfer gap is largest for paradigm tasks (30.7\%) and smallest for the Other category (9.6\%), suggesting that pattern-based tasks benefit most from slides but also show the sharpest drop-off when generalization is required. Note that Morphophonology ($n = 10$) and Other ($n = 12$, merging four subtypes) have small sample sizes; conclusions for these categories are exploratory.

\begin{table}[t]
\small
\centering
\setlength{\tabcolsep}{3pt}
\begin{tabular}{l|cccc}
\toprule
\textbf{System} & \textbf{Parad.} & \textbf{Trans.} & \textbf{Morph.} & \textbf{Other} \\
& \textit{(n=37)} & \textit{(n=31)} & \textit{(n=10)} & \textit{(n=12)} \\
\midrule
Claude Opus 4.7 & \gup{8.5} & \gup{0.8} & \gup{2.2} & \gup{11.1} \\
Qwen3.6-Max & \gup{9.4} & \gup{0.5} & \gup{2.1} & \gup{6.2} \\
MiniMax-M2.7 & \gup{7.9} & \gdn{0.9} & \gdn{2.2} & \gup{4.4} \\
GPT-5.4 & \gup{4.5} & \gdn{0.4} & \gup{2.3} & \gup{4.9} \\
Claude Sonnet 4.6 & \gup{4.7} & \gdn{0.8} & \gdn{2.3} & \gup{3.3} \\
Qwen3.6-Plus & \gup{3.0} & \gdn{1.4} & \gdn{5.6} & \gup{1.8} \\
Kimi-K2.6 & \gdn{0.0} & \gdn{1.4} & \gdn{2.1} & \gdn{1.2} \\
GLM-5.1 & \gup{0.8} & \gdn{3.6} & \gdn{2.1} & \gdn{0.9} \\
DeepSeek-V4-Pro & \gdn{1.2} & \gdn{2.5} & \gdn{2.4} & \gdn{1.9} \\
Gemini-3.1-Pro & \gdn{2.9} & \gdn{1.1} & \gdn{5.9} & \gdn{2.8} \\
\midrule
Mean & \gup{3.5} & \gdn{1.1} & \gdn{1.6} & \gup{2.5} \\
\bottomrule
\end{tabular}
\caption{\textbf{D4 gain (\%) by task type.} Paradigm completion is the most teachable; translation tasks show weakest gains.}
\label{tab:tasktype_results}
\end{table}

\begin{figure}[t]
\centering
\includegraphics[width=\columnwidth]{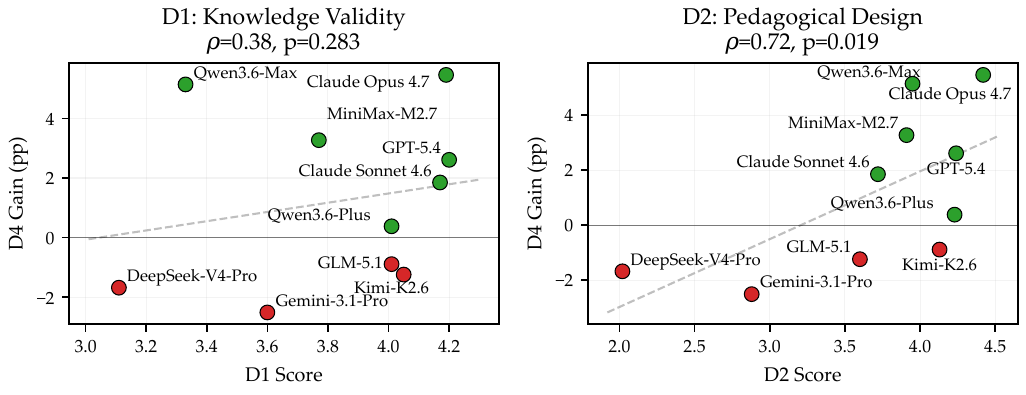}
\caption{D1/D2 scores vs.\ D4 gain. D2 (pedagogical design) shows a stronger positive association with learning gain ($\rho = 0.72$) than D1 (knowledge validity; $\rho = 0.38$), though $N = 10$ limits statistical power. Green = positive gain; red = negative gain.}
\label{fig:d1d2_vs_gain}
\vspace{-1em}
\end{figure}

\paragraph{Artifact quality vs.\ learning outcomes.}
To quantify the relationship between artifact quality and instructional effectiveness, we compute Spearman rank correlations between D1/D2 scores and D4 gain across 10 systems (Figure~\ref{fig:d1d2_vs_gain}). Within the evaluated systems, D2 (pedagogical design) shows a stronger positive association with gain ($\rho = 0.72$, $p = 0.019$) than D1 (knowledge validity; $\rho = 0.38$, $p = 0.28$). However, with only $N = 10$ systems, confidence intervals are wide and individual outliers can substantially shift these estimates; these correlations should be interpreted as suggestive rather than conclusive. The qualitative pattern is nonetheless informative: GPT-5.4 scores highest on D1 (4.20) yet ranks only fourth on gain, while MiniMax-M2.7 scores below average on D1 (3.77) yet ranks third, illustrating that content validity alone is not sufficient for effective instruction.

\paragraph{Case study.}
Per-puzzle analysis reveals extreme variation: the best single-puzzle gain is \gup{68}\% (Wik-Mungkan compound nouns, Claude Opus 4.7), while the worst is \gdn{50}\% (Jahai smell verb semantics, Gemini-3.1-Pro). In the success case, a low-pretest puzzle (4\%) reaches 72\% after studying a deck with clear tabular rule presentation. In the failure case, a high-pretest puzzle (72\%) drops to 22\%, suggesting the deck disrupts useful prior knowledge. We also identify an artifact-quality gap case: GPT-5.4 achieves perfect near-transfer (100\%) on a Dutch spelling puzzle yet causes a 30\% posttest decline, demonstrating that visual polish and local pattern learning do not guarantee effective instruction. Detailed analysis with representative slide excerpts is provided in Appendix~\ref{app:case_study}.

\section{Discussion}
\label{sec:discussion}

\subsection{Evaluation Reliability}
\label{sec:reliability}

Table~\ref{tab:reliability-slate} reports agreement statistics for D1 and D2. Human--human ICC reaches 0.80 for D1 and 0.72 for D2, indicating excellent and good reliability respectively. Human--model Pearson $r$ reaches 0.80 for D1 and 0.69 for D2. Agreement is stronger on objective content-validity sub-dimensions (e.g., rule coverage) and more moderate on subjective pedagogical sub-dimensions (e.g., scaffolding quality). We observe that model judges tend to over-score D2 when decks exhibit surface structural completeness without genuine pedagogical depth, motivating the use of multi-agent panels.

\begin{table}[t]
    \centering
    \small
    \setlength{\tabcolsep}{3pt}
    \begin{tabular}{l@{\hspace{6pt}}c@{\hspace{6pt}}c}
        \toprule
        \textbf{Metric} & \textbf{H--H} & \textbf{H--M} \\
        & \textbf{ICC} & \textbf{Pearson $r$} \\
        \midrule
        D1.1 Rule coverage & 0.82 & 0.81 \\
        D1.2 Data fidelity & 0.78 & 0.79 \\
        D1.3 Metalinguistic acc. & 0.71 & 0.75 \\
        D1.4 Answer non-leakage & 0.85 & 0.83 \\
        D1.5 Scope approp. & 0.76 & 0.77 \\
        \midrule
        D1 average & 0.80 & 0.80 \\
        \midrule
        D2.1 Cognitive load & 0.73 & 0.72 \\
        D2.2 Worked examples & 0.68 & 0.68 \\
        D2.3 Scaffolding & 0.65 & 0.65 \\
        D2.4 Dual-channel integ. & 0.70 & 0.70 \\
        D2.5 Practice quality & 0.58 & 0.51 \\
        \midrule
        D2 average & 0.72 & 0.69 \\
        \bottomrule
    \end{tabular}
    \caption{D1/D2 reliability. H--H: human--human ICC; H--M: human--model Pearson $r$.}
    \label{tab:reliability-slate}
    \vspace{-1.2em}
\end{table}

\subsection{Construct Validity: Error Injection}

The error injection study (Appendix~\ref{app:error_injection}) provides construct-oriented validation by selectively corrupting strong reference decks along four failure families. Each perturbation degrades its target dimension most severely: knowledge corruption drops D1 by 4.5 points (on a 10-point scale) while leaving D2/D3 nearly intact; pedagogical corruption drops D2 by 4.4 points; and visual corruption drops D3 by 5.3 points. Assessment-integrity corruption is diagnostically distinct: it inflates posttest scores by 0.5 points while destroying far-transfer performance ($-$5.0 points), confirming that answer leakage creates artificial gains without genuine learning. This selective sensitivity supports the claim that \textbf{\texttt{SLATE}} dimensions measure interpretable instructional properties rather than surface fluency.

\subsection{Failure Modes and Hypothesized Mechanisms}

Three recurring failure modes emerge. First, \emph{content-faithfulness drift}: decks appear fluent and pedagogically structured, yet examples or rule formulations contain errors (reflected in weak D1.2 data fidelity scores). This is supported by D1 sub-scores: rule coverage and scope score high across systems (surface-level completeness), but data fidelity is consistently weaker, as plausibility-optimizing models fail to catch constructed-example errors.

Second, \emph{pedagogical shallowness}: systems produce the form of a lesson (headings, worked examples, practice sections) without instructional coherence. D2 correlates far more strongly with D4 gain ($\rho = 0.72$) than D1 ($\rho = 0.38$), showing instructional organization matters more than content correctness alone, consistent with Cognitive Load Theory's emphasis on sequencing and scaffolding.

Third, \emph{near-transfer bias}: all systems support local pattern reuse far better than deeper abstraction, as shown by the persistent 20+\% near/far transfer gap. Following \citet{Barnett}, our far-transfer items require cross-domain generalization, which fails without explicit abstraction support. Current decks rarely provide the contrastive examples or analogical bridges needed for novel context application.

\subsection{VLM Student Validity}
\label{sec:VLM_Student_Validity}

Using a VLM as a learner proxy is scalable but raises validity concerns. The pretest baseline of 23.78\% likely reflects pretrained language exposure rather than domain-specific memorization, since the items require structured rule application. The ablation study validates that slide effects scale with model capacity (negative for 3B, positive for 8B+), consistent with research on the expertise reversal effect~\citep{Cognitive}.

To distinguish genuine learning from contextual priming, we conduct a content perturbation control: preserving slide structure and visual design but replacing teaching content with material from unrelated languages. Under this condition, posttest gains fall below the zero-information baseline, confirming that observed gains are content-driven. We further validate the VLM proxy through a three-system human pilot with 30 participants, covering high-gain Claude Opus 4.7~\citep{claudeOpus4.7}, intermediate-gain Claude Sonnet 4.6~\citep{claudeSonnet4.6}, and negative-gain Gemini-3.1-Pro~\citep{gemini31pro}. Human learners show a significant overall posttest gain ($t(29) = 6.78$, $p < 0.001$), with 85.7\% per-puzzle directional agreement and perfect cross-system ranking consistency (Spearman $\rho = 1.0$) between VLM and human results. Full experimental details, group-level results and statistical analyses are provided in Appendix~\ref{app:human_pilot} and Appendix~\ref{app:Human_Study}.

\subsection{Broader Implications}

The dissociation between artifact quality and instructional effectiveness has implications beyond slide generation. Any ``generation + evaluation'' benchmark that relies solely on artifact-side rubrics risks the same validity gap: materials that appear well-structured may not function as intended when deployed. This parallels findings in the LLM-as-judge literature~\citep{Validates}, where judge scores correlate imperfectly with downstream outcomes. We note that \textbf{\texttt{SLATE}}'s current scope is limited to explicit rule-based instruction with discrete, teachable rules and clear input-output mappings. Real language teaching also involves context-dependent pragmatics and communicative interaction; extending the evaluation framework to these less rule-governed domains is an important direction for future work.

\section{Conclusion}

The central finding of this work is that artifact quality and instructional effectiveness are dissociated: AI-generated slides that appear pedagogically well-structured under rubric evaluation do not reliably produce learning gains or support knowledge transfer. Within the evaluated systems, pedagogical design (D2) shows a stronger association with learning outcomes than content validity (D1), suggesting that instructional organization may matter more than factual coverage alone. The persistent near/far transfer gap across all systems indicates that current generation approaches support pattern reuse but not the deeper abstraction needed for genuine conceptual generalization. A content perturbation control and a three-system human pilot (n=30) support the validity of the VLM-based evaluation protocol.

The most important next steps are scaling human validation and incorporating transfer-oriented objectives into generation pipelines. Extending \textbf{\texttt{SLATE}} to interactive, multi-session formats is a promising direction for closing the near/far transfer gap.

\section*{Limitations}

\paragraph{VLM vs.\ human learners.} D4 relies on a VLM student proxy whose cross-system ranking validity has been confirmed via a three-system human pilot (n=30) spanning the full range of instructional effectiveness observed in our experiments. While VLM learners differ from human learners in several aspects—including episodic memory, interactive question-asking ability, and static image processing—their perfect ranking consistency with human learners across tested systems supports their use for scalable cross-system comparison. Extending validation to additional model architectures and larger participant cohorts remains a valuable direction for future work.

\paragraph{Static, single-session evaluation.} \textbf{\texttt{SLATE}} evaluates one-shot, non-interactive study. Real instruction involves iterative feedback, learner questions, and spaced repetition, none of which are captured in the current protocol. Extending the benchmark to multi-turn and interactive formats is a natural next step.

\section*{Ethics Statement}

\textbf{\texttt{SLATE}} is an evaluation benchmark, not evidence that AI-generated teaching slides are ready for unsupervised educational deployment. Incorrect explanations or misleading visualizations may negatively affect learners. Human oversight remains necessary in real instructional settings.

\bibliography{main}

\begin{thebibliography}{51}
\providecommand{\natexlab}[1]{#1}

\bibitem[{{Anthropic}(2026{\natexlab{a}})}]{claudeOpus4.7}
{Anthropic}. 2026{\natexlab{a}}.
\newblock \href {https://www.anthropic.com/news/claude-opus-4-7} {Introducing claude opus 4.7}.

\bibitem[{{Anthropic}(2026{\natexlab{b}})}]{claudeSonnet4.6}
{Anthropic}. 2026{\natexlab{b}}.
\newblock \href {https://www.anthropic.com/claude-sonnet-4-6-system-card} {System card: Claude sonnet 4.6}.

\bibitem[{Bannò et~al.(2024)Bannò, Knill, and Gales}]{Grammatical}
Stefano Bannò, Kate Knill, and Mark J.~F. Gales. 2024.
\newblock \href {https://arxiv.org/abs/2408.09565} {Grammatical error feedback: An implicit evaluation approach}.
\newblock \emph{Preprint}, arXiv:2408.09565.

\bibitem[{Barnett and Ceci(2002)}]{Barnett}
Susan~M. Barnett and Stephen~J. Ceci. 2002.
\newblock \href {https://api.semanticscholar.org/CorpusID:3923174} {When and where do we apply what we learn? a taxonomy for far transfer.}
\newblock \emph{Psychological bulletin}, 128 4:612--37.

\bibitem[{Bean et~al.(2024)Bean, Hellsten, Mayne, Magomere, Chi, Chi, Hale, and Kirk}]{LINGOLY}
Andrew~M. Bean, Simi Hellsten, Harry Mayne, Jabez Magomere, Ethan~A. Chi, Ryan Chi, Scott~A. Hale, and Hannah~Rose Kirk. 2024.
\newblock \href {https://arxiv.org/abs/2406.06196} {Lingoly: A benchmark of olympiad-level linguistic reasoning puzzles in low-resource and extinct languages}.
\newblock \emph{Preprint}, arXiv:2406.06196.

\bibitem[{Chen et~al.(2026)Chen, Zhu, lin Li, Wang, Yang, and Guo}]{PresentBench}
Xin-Sheng Chen, Jiayu Zhu, Pei lin Li, Hanzheng Wang, Shuojin Yang, and Meng-Hao Guo. 2026.
\newblock \href {https://arxiv.org/abs/2603.07244} {Presentbench: A fine-grained rubric-based benchmark for slide generation}.
\newblock \emph{Preprint}, arXiv:2603.07244.

\bibitem[{Chi et~al.(2024)Chi, Malchev, Kong, Chi, Huang, Chi, McCoy, and Radev}]{modeLing}
Nathan~A. Chi, Teodor Malchev, Riley Kong, Ryan~A. Chi, Lucas Huang, Ethan~A. Chi, R.~Thomas McCoy, and Dragomir Radev. 2024.
\newblock \href {https://arxiv.org/abs/2406.17038} {modeling: A novel dataset for testing linguistic reasoning in language models}.
\newblock \emph{Preprint}, arXiv:2406.17038.

\bibitem[{Clark and Kimmons(2023)}]{CognitiveLoad}
Chad Clark and Royce Kimmons. 2023.
\newblock \href {https://doi.org/10.59668/371.12980} {Cognitive load theory}.
\newblock \emph{EdTechnica}.

\bibitem[{{DeepSeek-AI}(2026)}]{deepseek2026v4}
{DeepSeek-AI}. 2026.
\newblock \href {https://huggingface.co/deepseek-ai/DeepSeek-V4-Flash/blob/main/DeepSeek_V4.pdf} {{DeepSeek-V4}: Towards highly efficient million-token context intelligence}.

\bibitem[{Do et~al.(2024)Do, Shafqat, Ling, and Sarda}]{PAIGE}
Tiffany~D. Do, Usama~Bin Shafqat, Elsie Ling, and Nikhil Sarda. 2024.
\newblock \href {https://arxiv.org/abs/2409.04645} {Paige: Examining learning outcomes and experiences with personalized ai-generated educational podcasts}.
\newblock \emph{Preprint}, arXiv:2409.04645.

\bibitem[{Fu et~al.(2022)Fu, Wang, McDuff, and Song}]{DOC2PPT}
Tsu-Jui Fu, William~Yang Wang, Daniel McDuff, and Yale Song. 2022.
\newblock \href {https://arxiv.org/abs/2101.11796} {Doc2ppt: Automatic presentation slides generation from scientific documents}.
\newblock \emph{Preprint}, arXiv:2101.11796.

\bibitem[{Gao et~al.(2025)Gao, Liu, Yue, Yao, Lv, Zhang, Wang, and Huang}]{Agent4Edu}
Weibo Gao, Qi~Liu, Linan Yue, Fangzhou Yao, Rui Lv, Zheng Zhang, Hao Wang, and Zhenya Huang. 2025.
\newblock \href {https://arxiv.org/abs/2501.10332} {Agent4edu: Generating learner response data by generative agents for intelligent education systems}.
\newblock \emph{Preprint}, arXiv:2501.10332.

\bibitem[{Ge et~al.(2025)Ge, Wang, Zhou, Peng, Subramanian, Tan, Sap, Suhr, Fried, Neubig, and Darrell}]{AutoPresent}
Jiaxin Ge, Zora~Zhiruo Wang, Xuhui Zhou, Yi-Hao Peng, Sanjay Subramanian, Qinyue Tan, Maarten Sap, Alane Suhr, Daniel Fried, Graham Neubig, and Trevor Darrell. 2025.
\newblock \href {https://arxiv.org/abs/2501.00912} {Autopresent: Designing structured visuals from scratch}.
\newblock \emph{Preprint}, arXiv:2501.00912.

\bibitem[{Gehrmann et~al.(2022)Gehrmann, Clark, and Sellam}]{Repairing}
Sebastian Gehrmann, Elizabeth Clark, and Thibault Sellam. 2022.
\newblock \href {https://arxiv.org/abs/2202.06935} {Repairing the cracked foundation: A survey of obstacles in evaluation practices for generated text}.
\newblock \emph{Preprint}, arXiv:2202.06935.

\bibitem[{GLM-5-Team et~al.(2026)GLM-5-Team, :, Zeng, Lv, Hou, Du, Zheng, Chen, Yin, Ge, Huang, Xie, Zhu, Yin, Wang, Pan, Zeng, Zhang, Wang, Chen, Zhang, Jiao, Guo, Wang, Du, Wu, Wang, Li, Fan, Zhong, Liu, Zhao, Du, Dong, Lu, Shuang-Li, Cao, Liu, Jiang, Chen, Zhang, Huang, Dong, Xu, Wei, An, Niu, Zhu, Wen, Cen, Bai, Qiao, Wang, Wang, Zhu, Liu, Li, Wang, Wen, Huang, Cai, Yu, Li, Hu, Zhang, Zhang, Lin, Yang, Wang, Ai, Zhu, Yi, Chen, Wen, Sun, Zhao, Hu, Zhang, Liu, Zhang, Peng, Tai, Zhang, Liu, Wang, Yan, Ge, Liu, Chu, Zhao, Wang, Zhao, Ren, Wang, Zhang, Gui, Zhao, Li, An, Li, Yuan, Du, Liu, Zhi, Duan, Zhou, Wei, Wang, Luo, Zhang, Sha, Xu, Wu, Ding, Chen, Li, Lin, Ta, Zou, Song, Yang, Tu, Yang, Wu, Zhang, Li, Li, Fan, Qin, Tian, Zhang, Yu, Liang, Kuang, Cheng, Li, Yan, Hu, Ling, Fan, Xia, Zhang, Zhang, Pan, Zou, Zhang, Liu, Wu, Li, Wang, Zhu, Tan, Zhou, Pan, Zhang, Su, Geng, Yan, Tan, Bi, Shen, Yang, Li, Liu, Wang, Li, Wu, Zhang, Duan, Zhang, Liu, Jiang, Yan, Zhang, Wei, Chen, Feng, Yao, Chai, Wang, Zhang, Xu,
  Huang, Wang, Li, Dong, and Tang}]{glm5team2026glm5vibecodingagentic}
GLM-5-Team, :, Aohan Zeng, Xin Lv, Zhenyu Hou, Zhengxiao Du, Qinkai Zheng, Bin Chen, Da~Yin, Chendi Ge, Chenghua Huang, Chengxing Xie, Chenzheng Zhu, Congfeng Yin, Cunxiang Wang, Gengzheng Pan, Hao Zeng, Haoke Zhang, Haoran Wang, and 168 others. 2026.
\newblock \href {https://arxiv.org/abs/2602.15763} {Glm-5: from vibe coding to agentic engineering}.
\newblock \emph{Preprint}, arXiv:2602.15763.

\bibitem[{{Google DeepMind}(2026)}]{gemini31pro}
{Google DeepMind}. 2026.
\newblock \href {https://deepmind.google/models/gemini/pro/} {{Gemini 3.1 Pro}: Best for complex tasks and bringing creative concepts to life}.

\bibitem[{Hake(1998)}]{Interactive}
Richard Hake. 1998.
\newblock \href {https://doi.org/10.1119/1.18809} {Interactive-engagement versus traditional methods: A six-thousand-student survey of mechanics test data for introductory physics courses}.
\newblock \emph{American Journal of Physics - AMER J PHYS}, 66.

\bibitem[{Han and Choi(2025)}]{Beyond}
Junzhi Han and Jinho~D. Choi. 2025.
\newblock \href {https://doi.org/10.18653/v1/2025.bea-1.58} {Beyond linear digital reading: An {LLM}-powered concept mapping approach for reducing cognitive load}.
\newblock In \emph{Proceedings of the 20th Workshop on Innovative Use of NLP for Building Educational Applications (BEA 2025)}, pages 805--817, Vienna, Austria. Association for Computational Linguistics.

\bibitem[{Hendrycks et~al.(2021)Hendrycks, Burns, Basart, Zou, Mazeika, Song, and Steinhardt}]{Measuring}
Dan Hendrycks, Collin Burns, Steven Basart, Andy Zou, Mantas Mazeika, Dawn Song, and Jacob Steinhardt. 2021.
\newblock \href {https://arxiv.org/abs/2009.03300} {Measuring massive multitask language understanding}.
\newblock \emph{Preprint}, arXiv:2009.03300.

\bibitem[{Kasneci et~al.(2023)Kasneci, Sessler, Küchemann, Bannert, Dementieva, Fischer, Gasser, Groh, Günnemann, Hüllermeier, Krusche, Kutyniok, Michaeli, Nerdel, Pfeffer, Poquet, Sailer, Schmidt, Seidel, Stadler, Weller, Kuhn, and Kasneci}]{ChatGPT}
Enkelejda Kasneci, Kathrin Sessler, Stefan Küchemann, Maria Bannert, Daryna Dementieva, Frank Fischer, Urs Gasser, Georg Groh, Stephan Günnemann, Eyke Hüllermeier, Stephan Krusche, Gitta Kutyniok, Tilman Michaeli, Claudia Nerdel, Jürgen Pfeffer, Oleksandra Poquet, Michael Sailer, Albrecht Schmidt, Tina Seidel, and 4 others. 2023.
\newblock \href {https://doi.org/10.1016/j.lindif.2023.102274} {Chatgpt for good? on opportunities and challenges of large language models for education}.
\newblock \emph{Learning and Individual Differences}, 103:102274.

\bibitem[{Liang et~al.(2025)Liang, Zhang, Xu, Sun, and You}]{SlideGen}
Xin Liang, Xiang Zhang, Yiwei Xu, Siqi Sun, and Chenyu You. 2025.
\newblock \href {https://arxiv.org/abs/2512.04529} {Slidegen: Collaborative multimodal agents for scientific slide generation}.
\newblock \emph{Preprint}, arXiv:2512.04529.

\bibitem[{Lu and Wang(2024)}]{Generative}
Xinyi Lu and Xu~Wang. 2024.
\newblock \href {https://doi.org/10.1145/3657604.3662031} {Generative students: Using llm-simulated student profiles to support question item evaluation}.
\newblock In \emph{Proceedings of the Eleventh ACM Conference on Learning @ Scale}, page 16–27. ACM.

\bibitem[{Macina et~al.(2025)Macina, Daheim, Hakimi, Kapur, Gurevych, and Sachan}]{MathTutorBench}
Jakub Macina, Nico Daheim, Ido Hakimi, Manu Kapur, Iryna Gurevych, and Mrinmaya Sachan. 2025.
\newblock \href {https://arxiv.org/abs/2502.18940} {Mathtutorbench: A benchmark for measuring open-ended pedagogical capabilities of llm tutors}.
\newblock \emph{Preprint}, arXiv:2502.18940.

\bibitem[{Maurya et~al.(2025)Maurya, Srivatsa, Petukhova, and Kochmar}]{Unifying}
Kaushal~Kumar Maurya, Kv~Aditya Srivatsa, Kseniia Petukhova, and Ekaterina Kochmar. 2025.
\newblock \href {https://doi.org/10.18653/v1/2025.naacl-long.57} {Unifying {AI} tutor evaluation: An evaluation taxonomy for pedagogical ability assessment of {LLM}-powered {AI} tutors}.
\newblock In \emph{Proceedings of the 2025 Conference of the Nations of the Americas Chapter of the Association for Computational Linguistics: Human Language Technologies (Volume 1: Long Papers)}, pages 1234--1251, Albuquerque, New Mexico. Association for Computational Linguistics.

\bibitem[{Mayer(2002)}]{Multimedia}
Richard~E. Mayer. 2002.
\newblock \href {https://doi.org/10.1016/S0079-7421(02)80005-6} {Multimedia learning}.
\newblock volume~41 of \emph{Psychology of Learning and Motivation}, pages 85--139. Academic Press.

\bibitem[{Mayer(2005)}]{Mayer_2005}
Richard~E. Mayer. 2005.
\newblock Cognitive theory of multimedia learning.
\newblock In Richard~E. Mayer, editor, \emph{The Cambridge Handbook of Multimedia Learning}, pages 31--48. Cambridge University Press, Cambridge, UK.

\bibitem[{{MiniMax}(2026)}]{minimax2026m27}
{MiniMax}. 2026.
\newblock \href {https://github.com/MiniMax-AI/MiniMax-M2} {{MiniMax-M2.7}}.

\bibitem[{{Moonshot AI}(2026)}]{kimik26}
{Moonshot AI}. 2026.
\newblock \href {https://www.kimi.com/blog/kimi-k2-6} {{Kimi K2.6}: Advancing open-source coding}.

\bibitem[{Morris et~al.(1977)Morris, Bransford, and Franks}]{Levels}
C.~Donald Morris, John~D. Bransford, and Jeffery~J. Franks. 1977.
\newblock \href {https://doi.org/10.1016/S0022-5371(77)80016-9} {Levels of processing versus transfer appropriate processing}.
\newblock \emph{Journal of Verbal Learning and Verbal Behavior}, 16(5):519--533.

\bibitem[{Norris and Ortega(2002)}]{Effectiveness}
John Norris and Lourdes Ortega. 2002.
\newblock \href {https://doi.org/10.1111/0023-8333.00136} {Effectiveness of l2 instruction: A research synthesis and quantitative meta-analysis}.
\newblock \emph{Language Learning}, 50:417 -- 528.

\bibitem[{{OpenAI}(2026)}]{openai2026gpt54}
{OpenAI}. 2026.
\newblock \href {https://openai.com/index/introducing-gpt-5-4/} {Introducing gpt‑5.4}.

\bibitem[{Perkins and Salomon(1999)}]{Transfer}
David Perkins and Gavriel Salomon. 1999.
\newblock Transfer of learning.
\newblock 11.

\bibitem[{{Qwen Team}(2026)}]{qwen2026qwen36}
{Qwen Team}. 2026.
\newblock \href {https://github.com/QwenLM/Qwen3.6} {{Qwen3.6}: Large language model series}.

\bibitem[{SCHMIDT(1990)}]{Role}
RICHARD~W. SCHMIDT. 1990.
\newblock \href {https://doi.org/10.1093/applin/11.2.129} {The role of consciousness in second language learning1}.
\newblock \emph{Applied Linguistics}, 11(2):129--158.

\bibitem[{Shankar et~al.(2024)Shankar, Zamfirescu-Pereira, Hartmann, Parameswaran, and Arawjo}]{Validates}
Shreya Shankar, J.~D. Zamfirescu-Pereira, Björn Hartmann, Aditya~G. Parameswaran, and Ian Arawjo. 2024.
\newblock \href {https://arxiv.org/abs/2404.12272} {Who validates the validators? aligning llm-assisted evaluation of llm outputs with human preferences}.
\newblock \emph{Preprint}, arXiv:2404.12272.

\bibitem[{Shi et~al.(2025)Shi, Liang, and Xu}]{EducationQ}
Yao Shi, Rongkeng Liang, and Yong Xu. 2025.
\newblock \href {https://doi.org/10.18653/v1/2025.acl-long.1576} {Educationq: Evaluating llms’ teaching capabilities through multi-agent dialogue framework}.
\newblock In \emph{Proceedings of the 63rd Annual Meeting of the Association for Computational Linguistics (Volume 1: Long Papers)}, page 32799–32828. Association for Computational Linguistics.

\bibitem[{Srinivasa et~al.(2025)Srinivasa, Che, Zhang, Mares, Hernandez, Park, Lee, Mangialardi, Ng, Cardona, Gunjal, He, Liu, and Xing}]{TutorBench}
Rakshith~S Srinivasa, Zora Che, Chen Bo~Calvin Zhang, Diego Mares, Ernesto Hernandez, Jayeon Park, Dean Lee, Guillermo Mangialardi, Charmaine Ng, Ed-Yeremai~Hernandez Cardona, Anisha Gunjal, Yunzhong He, Bing Liu, and Chen Xing. 2025.
\newblock \href {https://arxiv.org/abs/2510.02663} {Tutorbench: A benchmark to assess tutoring capabilities of large language models}.
\newblock \emph{Preprint}, arXiv:2510.02663.

\bibitem[{Sweller(1988)}]{Cognitive}
John Sweller. 1988.
\newblock \href {https://doi.org/10.1016/0364-0213(88)90023-7} {Cognitive load during problem solving: Effects on learning}.
\newblock \emph{Cognitive Science}, 12(2):257--285.

\bibitem[{Sweller et~al.(2011)Sweller, Ayres, Kalyuga, and Chandler}]{Expertise}
John Sweller, Paul Ayres, Slava Kalyuga, and Paul Chandler. 2011.
\newblock \href {https://doi.org/10.1007/978-1-4419-8126-4_12} {The expertise reversal effect}.
\newblock \emph{Faculty of Education - Papers}, 38.

\bibitem[{Xie et~al.(2025)Xie, Waterfield, Kennedy, and Zhang}]{SlideBot}
Eric Xie, Danielle Waterfield, Michael Kennedy, and Aidong Zhang. 2025.
\newblock \href {https://arxiv.org/abs/2511.09804} {Slidebot: A multi-agent framework for generating informative, reliable, multi-modal presentations}.
\newblock \emph{Preprint}, arXiv:2511.09804.

\bibitem[{Xu et~al.(2025)Xu, Wen, Pan, Dominguez, Hu, and Zhang}]{Classroom}
Songlin Xu, Hao-Ning Wen, Hongyi Pan, Dallas Dominguez, Dongyin Hu, and Xinyu Zhang. 2025.
\newblock \href {https://doi.org/10.1145/3706598.3713773} {Classroom simulacra: Building contextual student generative agents in online education for learning behavioral simulation}.
\newblock In \emph{Proceedings of the 2025 CHI Conference on Human Factors in Computing Systems}, page 1–26. ACM.

\bibitem[{Yang et~al.(2026)Yang, Li, Ren, Lu, Wang, Huang, Zong, Zhan, and Li}]{SlidesGen-Bench}
Yunqiao Yang, Wenbo Li, Houxing Ren, Zimu Lu, Ke~Wang, Zhiyuan Huang, Zhuofan Zong, Mingjie Zhan, and Hongsheng Li. 2026.
\newblock {SlidesGen-Bench}: Evaluating slides generation via computational and quantitative metrics.
\newblock \emph{arXiv preprint arXiv:2601.09487}.

\bibitem[{Yuan et~al.(2026)Yuan, Xiao, Li, Xuan, Tong, Diab, and Mitchell}]{Towards}
Zhihao Yuan, Yunze Xiao, Ming Li, Weihao Xuan, Richard Tong, Mona Diab, and Tom Mitchell. 2026.
\newblock \href {https://arxiv.org/abs/2601.05473} {Towards valid student simulation with large language models}.
\newblock \emph{Preprint}, arXiv:2601.05473.

\bibitem[{Yue et~al.(2025)Yue, Lyu, Suh, Zhang, and Yao}]{MathVC}
Murong Yue, Wenhan Lyu, Jennifer Suh, Yixuan Zhang, and Ziyu Yao. 2025.
\newblock \href {https://arxiv.org/abs/2404.06711} {Mathvc: An llm-simulated multi-character virtual classroom for mathematics education}.
\newblock \emph{Preprint}, arXiv:2404.06711.

\bibitem[{Zeng et~al.(2025)Zeng, Ouyang, Cui, and Ng}]{slidetailor}
Wenzheng Zeng, Mingyu Ouyang, Langyuan Cui, and Hwee~Tou Ng. 2025.
\newblock \href {https://arxiv.org/abs/2512.20292} {Slidetailor: Personalized presentation slide generation for scientific papers}.
\newblock \emph{Preprint}, arXiv:2512.20292.

\bibitem[{Zhang et~al.(2026)Zhang, Li, Li, Guo, Wu, Zheng, Yang, Zhang, Li, Yan et~al.}]{realchart2code}
Jiajun Zhang, Yuying Li, Zhixun Li, Xingyu Guo, Jingzhuo Wu, Leqi Zheng, Yiran Yang, Jianke Zhang, Qingbin Li, Shannan Yan, and 1 others. 2026.
\newblock Realchart2code: Advancing chart-to-code generation with real data and multi-task evaluation.
\newblock \emph{arXiv preprint arXiv:2603.25804}.

\bibitem[{Zheng et~al.(2025{\natexlab{a}})Zheng, Guan, Kong, Zheng, Zhou, Lin, Lu, He, Han, and Sun}]{PPTAgent}
Hao Zheng, Xinyan Guan, Hao Kong, Jia Zheng, Weixiang Zhou, Hongyu Lin, Yaojie Lu, Ben He, Xianpei Han, and Le~Sun. 2025{\natexlab{a}}.
\newblock \href {https://arxiv.org/abs/2501.03936} {Pptagent: Generating and evaluating presentations beyond text-to-slides}.
\newblock \emph{Preprint}, arXiv:2501.03936.

\bibitem[{Zheng et~al.(2025{\natexlab{b}})Zheng, Guan, Kong, Zheng, Zhou, Lin, Lu, He, Han, and Sun}]{PPTEval}
Hao Zheng, Xinyan Guan, Hao Kong, Jia Zheng, Weixiang Zhou, Hongyu Lin, Yaojie Lu, Ben He, Xianpei Han, and Le~Sun. 2025{\natexlab{b}}.
\newblock \href {https://arxiv.org/abs/2501.03936} {Pptagent: Generating and evaluating presentations beyond text-to-slides}.
\newblock \emph{Preprint}, arXiv:2501.03936.

\bibitem[{Zheng et~al.(2023)Zheng, Chiang, Sheng, Zhuang, Wu, Zhuang, Lin, Li, Li, Xing, Zhang, Gonzalez, and Stoica}]{Judging}
Lianmin Zheng, Wei-Lin Chiang, Ying Sheng, Siyuan Zhuang, Zhanghao Wu, Yonghao Zhuang, Zi~Lin, Zhuohan Li, Dacheng Li, Eric~P. Xing, Hao Zhang, Joseph~E. Gonzalez, and Ion Stoica. 2023.
\newblock \href {https://arxiv.org/abs/2306.05685} {Judging llm-as-a-judge with mt-bench and chatbot arena}.
\newblock \emph{Preprint}, arXiv:2306.05685.

\bibitem[{Zhong et~al.(2023)Zhong, Cui, Guo, Liang, Lu, Wang, Saied, Chen, and Duan}]{AGIEval}
Wanjun Zhong, Ruixiang Cui, Yiduo Guo, Yaobo Liang, Shuai Lu, Yanlin Wang, Amin Saied, Weizhu Chen, and Nan Duan. 2023.
\newblock \href {https://arxiv.org/abs/2304.06364} {Agieval: A human-centric benchmark for evaluating foundation models}.
\newblock \emph{Preprint}, arXiv:2304.06364.

\bibitem[{Şahin et~al.(2020)Şahin, Kementchedjhieva, Rust, and Gurevych}]{PuzzLingMachines}
Gözde~Gül Şahin, Yova Kementchedjhieva, Phillip Rust, and Iryna Gurevych. 2020.
\newblock \href {https://arxiv.org/abs/2004.13161} {Puzzling machines: A challenge on learning from small data}.
\newblock \emph{Preprint}, arXiv:2004.13161.

\end{thebibliography}
\clearpage
\appendix

\renewcommand*\footnoterule{} 

\newcommand\blfootnote[1]{%
  \begingroup
  \renewcommand\thefootnote{}\footnote{#1}%
  \addtocounter{footnote}{-1}%
  \endgroup
}

\nolinenumbers
\section*{Appendix}
\startcontents[sections]
\printcontents[sections]{l}{1}{\setcounter{tocdepth}{2}}

\clearpage
\section{Appendix: Additional Related Work}
\label{app:Additional_Related_Work}

\subsection{LLM-as-Judge and Multi-Model Evaluation}
LLM judges can approximate human preference rankings, with multi-judge aggregation reducing single-model biases~\citep{Judging}. 
However, \citet{Validates} demonstrate that LLM judges diverge from human raters when evaluation requires genuine expertise rather than surface fluency assessment, and \citet{Repairing} recommend multi-faceted designs that combine automatic scoring with outcome-based measures. 
These findings motivate SLATE's use of a multi-agent panel for D1/D2 scoring alongside learner-side transfer measurement (D4) as an external validity anchor.

\subsection{Simulated Learners and VLM Proxies}
LLM-simulated student profiles can predict item difficulty and discriminability in a manner consistent with human response distributions~\citep{Generative}, and classroom-level generative agents reproduce established learning phenomena including spacing effects~\citep{Classroom}. 
A critical limitation is raised by \citet{Towards}, who show that apparent VLM performance improvements after instruction may reflect pretrained recall rather than genuine learning from the presented material. 
SLATE addresses this threat by using low-resource languages with negligible web presence and a strict pretest--posttest design to distinguish true learning from prior exposure.

\subsection{Multimedia Learning and Visual Communication}
Mayer's Cognitive Theory of Multimedia Learning~\citep{Mayer_2005, Multimedia} operationalizes dual-channel processing within an instructional design framework, deriving the signaling, contiguity, and coherence principles from controlled experiments. 
Recent work extends these principles to AI-generated content: \citet{Beyond} show that LLM-powered concept mapping reduces extraneous cognitive load compared to linear digital text. 
These principles directly inform SLATE's D2 (cognitive-pedagogical design) and D3 (visual communication quality) dimensions.

\subsection{Cognitive Load and Pedagogical Sequencing}
Cognitive Load Theory distinguishes intrinsic load arising from material complexity, extraneous load arising from poor design, and germane load associated with schema formation~\citep{Cognitive}. 
The expertise reversal effect~\citep{CognitiveLoad,Expertise} predicts that instructional supports beneficial for novices become counterproductive as expertise grows, explaining why several SLATE decks produce negative gains at high pretest baselines. 
\citet{ChatGPT} further note that current LLMs tend to produce content that maximizes coverage rather than managing cognitive load, consistent with our finding that artifact-level completeness does not guarantee learning outcomes.

\subsection{Transfer of Learning and Assessment Design}
Barnett and Ceci's taxonomy~\citep{Barnett} distinguishes transfer along knowledge domain, physical context, temporal context, and functional context, clarifying that near and far transfer are endpoints of a continuum rather than binary categories. 
Transfer-Appropriate Processing~\citep{Levels} predicts that retrieval success depends on the match between encoding and retrieval processes rather than on memory strength alone, supporting SLATE's finding that format mismatch between study decks and test items can depress posttest performance. 
Pretest--posttest designs with normalized gain remain the standard approach for measuring learning in controlled settings~\citep{Interactive}.

\subsection{Language Acquisition and Explicit Rule Instruction}
Meta-analyses confirm that explicit instruction produces larger and more durable gains than implicit exposure for complex, rule-governed structures such as morphological paradigms and agreement patterns~\citep{Role}. 
Schmidt's Noticing Hypothesis~\citep{Effectiveness} establishes that conscious attention to linguistic form is a necessary condition for acquisition, motivating the explicit rule presentation required in SLATE's instructional units. 
However, \citet{Grammatical} find that LLM-generated grammatical feedback frequently lacks metalinguistic accuracy, underscoring the need for rigorous D1 (knowledge validity) evaluation of AI-generated teaching materials.

\section{Benchmark Details}
\label{app:benchmark_details}

\subsection{Instructional Unit Construction}

Each of the 90 source puzzles is transformed into one instructional unit. An instructional unit stores (i) the target knowledge point, (ii) extracted teachable rules, (iii) key terms, (iv) known exceptions, (v) representative original items, and (vi) transfer items. Across the benchmark, each unit contains an average of 5.8 teachable rules. These rules are intended to capture minimal explanatory generalizations rather than answer templates.

A teachable rule is defined as the smallest explanatory statement that a learner must acquire in order to solve the source puzzle and generalize beyond it. Depending on task type, such rules may describe morphological alternations, subject--agreement markers, translation correspondences, structural templates, ordering constraints, or exception conditions. This design ensures that slide generation targets an interpretable micro-curriculum rather than a direct answer key.

\subsection{Difficulty Rubric and Band Mapping}

Instructional difficulty is modeled using five 0--2 factors. The total score is the sum of these five factors, with a maximum of 10. Difficulty bands are assigned as follows: total scores of 0--3 are mapped to \textbf{Basic}, 4--6 to \textbf{Intermediate}, and 7--10 to \textbf{Challenge}.

\begin{table*}[h]
    \centering
    \small
    \setlength{\tabcolsep}{5pt}
    \begin{tabular}{l|p{0.18\textwidth}|p{0.18\textwidth}|p{0.18\textwidth}}
        \toprule
        \textbf{Factor} & \textbf{0 points} & \textbf{1 point} & \textbf{2 points} \\
        \midrule
        Rule complexity & Single rule with direct application & Two conditions or a limited rule variant & Interacting conditions, chained rules, or multiple dependent transformations \\
        Exception density & No explicit exceptions & One or two regularized exceptions & Three or more exceptions, or exceptions that themselves require rule learning \\
        Representation density & Compact material, low symbolic load & Moderate data length or moderate symbolic complexity & Large representations, dense glossing, IPA / tone-heavy notation, or high symbolic compression \\
        Transfer abstraction & Transfer closely resembles source application & Transfer requires modest recombination or cross-example abstraction & Transfer requires abstraction beyond surface format or combination of multiple rules \\
        Distractor strength & No meaningful distractors & Limited surface-similar distractors & Multiple strong distractors that align with plausible learner misconceptions \\
        \bottomrule
    \end{tabular}
    \caption{Instructional difficulty rubric used to assign Basic / Intermediate / Challenge bands.}
    \label{tab:difficulty_rubric}
\end{table*}

This rubric models how difficult a concept is to teach rather than merely how difficult the original puzzle is to solve. A problem may be challenging to infer from raw evidence but comparatively easy to explain once the core rule is known; conversely, a short-answer puzzle may still require a dense multi-step lesson if the rule system is heavily conditioned.

\subsection{Representative Examples of Teachable Rules}

The average of 5.8 teachable rules per unit is not meant to imply that all units are equally fragmented. Instead, it reflects the fact that even compact linguistics-olympiad lessons often require a small explanatory system rather than one single statement. Representative examples include the following:

\begin{itemize}
    \item \textbf{Agreement rule:} ``The prefix \texttt{o-} marks second-person plural subject on verbs.''
    \item \textbf{Conditional template:} ``Conditional clauses use the frame \texttt{inapa [condition], apo [result]}.''
    \item \textbf{Morphological boundary rule:} ``The suffix \texttt{-hi} marks plurality, while \texttt{-na} marks possession.''
    \item \textbf{Negation placement rule:} ``The negation marker \texttt{ega} precedes the constituent it negates.''
    \item \textbf{Exception boundary rule:} ``The connective \texttt{babana} introduces a reason relation rather than a conditional frame, and should not be generalized as a conditional marker.''
\end{itemize}

These examples illustrate why the benchmark requires a structured course outline and not only a compressed textual summary. A usable teaching deck must sequence such rules, relate them to evidence, and signal which ones are core versus exceptional.

\subsection{Course Outline Fields}

The course outline passed to the generation model contains five mandatory fields: \textbf{Learning Objective}, \textbf{Must-Cover Rules}, \textbf{Must-Not-Include}, \textbf{Visual Aid Required}, and \textbf{Transfer Design}. Together, these fields define what the deck must teach, what it must avoid, and how it should support later transfer.

\paragraph{Learning Objective.}
This field defines the target knowledge point in teachable terms. It is intentionally phrased as a lesson objective rather than as a puzzle-solving goal.

\paragraph{Must-Cover Rules.}
This field lists the rules, contrasts, and exceptions that the deck is required to explain. It anchors D1 and constrains the content space of the generator.

\paragraph{Must-Not-Include.}
This field explicitly prohibits direct answer exposure, irrelevant digressions, and content that would compromise evaluation integrity.

\paragraph{Visual Aid Required.}
This field specifies pedagogically necessary visual supports, such as aligned tables, gloss layouts, color-coded contrasts, arrows, segmentation markers, or boundary highlighting.

\paragraph{Transfer Design.}
This field specifies the intended generalization target and the kinds of examples that should help the learner bridge from source items to near and far transfer.

A crucial integrity constraint is that transfer answers are never exposed to the generator. The outline may require support for transfer, but it does not reveal the transfer solutions.

\subsection{Error Injection Design}

To validate whether the benchmark dimensions respond to the kinds of instructional defects they are intended to measure, we design a controlled perturbation setting based on four error families: knowledge errors, pedagogical errors, visual errors, and assessment-integrity errors. The purpose of this setting is not merely to create synthetic ablations, but to provide construct-oriented evidence that the benchmark is sensitive to interpretable educational failures.

\paragraph{Knowledge Errors.}
These perturbations alter one or more target rules, example labels, glosses, or exception boundaries while preserving surface fluency. They test whether D1 is sensitive to content corruption and whether D4 degrades when the learner is taught the wrong generalization.

\paragraph{Pedagogical Errors.}
These perturbations break the intended teaching logic. Concretely, they may remove worked examples, invert the example-to-rule progression, overload a page with dense unsupported rule statements, or mismatch practice items to what has been taught. In this benchmark, pedagogical error is therefore defined negatively as violation of the benchmark's own teaching criteria: objective transparency, scaffolded progression, manageable cognitive load, misconception prevention, and generative practice.

\paragraph{Visual Errors.}
These perturbations remove signals, disrupt visual consistency, or replace interpretable structures with dense plain text. Typical examples include removing alignment tables, deleting highlighting, mixing visual codes across pages, or flattening a rule visualization into prose.

\paragraph{Assessment-Integrity Errors.}
These perturbations introduce answer leakage or expose information that can artificially improve post-test performance without genuine teaching. This category is particularly important because a deck can appear ``effective'' under D4 for the wrong reason if evaluation integrity is not preserved.

\subsection{Error Injection Protocol, Expected Results, and Research Value}

A minimal solid design uses a paired comparison protocol. We begin with a set of high-quality decks, then create one controlled corruption per error family while keeping the rest of the deck fixed as much as possible. This produces paired variants such as \textit{original}, \textit{knowledge-corrupted}, \textit{pedagogically-corrupted}, \textit{visually-corrupted}, and \textit{leakage-corrupted}. The benchmark is then re-run on these paired variants.

The expected pattern is not that every metric drops uniformly, but that the drops are structured. Knowledge corruption should primarily depress D1 and then weaken D4 through mislearning. Pedagogical corruption should primarily depress D2 and often reduce post-test gain and near transfer. Visual corruption should primarily depress D3 and may reduce post-test gain by making the rule harder to decode. Corruption of exception handling and abstraction support should be especially damaging to far transfer. Leakage corruption is different: it may inflate post-test scores while simultaneously invalidating D1 and weakening the interpretability of D4. This makes it a particularly important diagnostic category.

The scientific value of this design is twofold. First, it provides construct validation: if the benchmark dimensions are meaningful, they should respond selectively to the corresponding perturbations. Second, it provides a diagnostic lens for future systems: rather than only reporting that one deck scores lower than another, researchers can test which kind of instructional failure a system is most vulnerable to.

\section{Theory-to-Metric Mapping}
\label{app:theory_mapping}

The evaluation framework is grounded in a set of learning-science theories rather than in a purely empirical checklist. This appendix summarizes the main mapping from theory to dimensions.

\begin{table*}[h]
    \centering
    \small
    \setlength{\tabcolsep}{5pt}
    \begin{tabular}{l|p{0.23\textwidth}|p{0.38\textwidth}}
        \toprule
        \textbf{Theory} & \textbf{Primary dimensions} & \textbf{Metric implications} \\
        \midrule
        Cognitive Load Theory & D2, D3 & Supports metrics for information density, load control, and reduction of extraneous complexity \\
        Multimedia Learning Theory & D2, D3 & Supports signaling, segmentation, layout coherence, and alignment between explanation and example \\
        Dual Coding Theory & D3 & Supports verbal--visual pairing, rule visualization, and interpretable non-verbal encoding \\
        Scaffolding / ZPD & D2 & Supports progressive sequencing, guided explanation, and the transition from supported to more independent application \\
        Transfer Theory & D4 & Supports the distinction between near and far transfer and motivates abstraction-sensitive evaluation \\
        Signaling / Communication Theory & D3, D2 & Supports consistent use of emphasis, highlighting, grouping, and other cues that guide interpretation \\
        Rule-Induction / SLA perspectives & D1, D4 & Supports correctness, exception handling, exemplar adequacy, and boundary-sensitive transfer \\
        \bottomrule
    \end{tabular}
    \caption{High-level mapping from learning-science theory to the four benchmark dimensions.}
    \label{tab:theory_mapping}
\end{table*}

These theories are used to justify not only what is measured, but why those measurements should matter for instruction. In particular, the framework distinguishes between content correctness (D1), teaching process quality (D2), visual support for interpretation (D3), and actual learner change (D4), rather than collapsing them into a single document-quality judgment.

\section{Evaluation Details}
\label{app:evaluation_details}

\subsection{Theory-Grounded Metrics}

The evaluation framework is grounded in cognitive load theory, multimedia learning theory, dual coding theory, scaffolding theory, transfer theory, and signaling theory. D1 focuses on linguistic knowledge validity; D2 on cognitive-pedagogical design; D3 on visual communication quality; and D4 on learning and transfer effectiveness.

\subsection{Human Expert Verification for D1 / D2}

The main D1 / D2 scoring pipeline uses expert-authored rubrics together with scalable model-assisted scoring. To validate this setup, we conducted a dedicated human expert verification study.

\paragraph{Experts.}
Three human linguistics experts participated in the verification stage. All held doctoral training in linguistics or applied linguistics. Two had research experience in low-resource languages, and one had more than three years of language teaching and slide-design experience. All experts were blind to system identity and had not participated in slide generation or prior machine scoring.

\paragraph{Sampling and Rating Procedure.}
We sampled 30 decks using stratified random sampling over difficulty band and system identity. After quality control, 29 decks remained valid. Experts first completed rubric calibration on five non-sampled decks and then independently rated the sampled decks on the five D1 sub-dimensions and five D2 sub-dimensions.

\subsection{Agreement Results by Sub-dimension}

\begin{table*}[h]
    \centering
    \small
    \setlength{\tabcolsep}{4pt}
    \begin{tabular}{l|lcc}
        \toprule
        \textbf{Dimension} & \textbf{Sub-dimension} & \textbf{ICC} & \textbf{Agreement level} \\
        \midrule
        D1 & D1.1 Rule coverage & 0.82 & Excellent \\
        D1 & D1.2 Data fidelity & 0.78 & Excellent \\
        D1 & D1.3 Metalinguistic accuracy & 0.71 & Good \\
        D1 & D1.4 Answer non-leakage & 0.85 & Excellent \\
        D1 & D1.5 Scope appropriateness & 0.76 & Good \\
        D1 & D1 average & 0.80 & Excellent \\
        \midrule
        D2 & D2.1 Cognitive load management & 0.73 & Good \\
        D2 & D2.2 Worked examples & 0.68 & Good \\
        D2 & D2.3 Scaffolding progression & 0.65 & Good \\
        D2 & D2.4 Dual-channel integration & 0.70 & Good \\
        D2 & D2.5 Practice opportunity & 0.58 & Moderate \\
        D2 & D2 average & 0.72 & Good \\
        \bottomrule
    \end{tabular}
    \caption{Human--human agreement for D1 and D2 using intraclass correlation coefficients.}
    \label{tab:human_icc}
\end{table*}

\begin{table*}[h]
    \centering
    \small
    \setlength{\tabcolsep}{4pt}
    \begin{tabular}{l|lcc}
        \toprule
        \textbf{Dimension} & \textbf{Sub-dimension} & \textbf{Pearson $r$} & \textbf{MAE} \\
        \midrule
        D1 & D1.1 Rule coverage & 0.81 & 0.32 \\
        D1 & D1.2 Data fidelity & 0.79 & 0.35 \\
        D1 & D1.3 Metalinguistic accuracy & 0.75 & 0.41 \\
        D1 & D1.4 Answer non-leakage & 0.83 & 0.28 \\
        D1 & D1.5 Scope appropriateness & 0.77 & 0.38 \\
        D1 & D1 average & 0.80 & 0.34 \\
        \midrule
        D2 & D2.1 Cognitive load management & 0.72 & 0.45 \\
        D2 & D2.2 Worked examples & 0.68 & 0.50 \\
        D2 & D2.3 Scaffolding progression & 0.65 & 0.52 \\
        D2 & D2.4 Dual-channel integration & 0.70 & 0.48 \\
        D2 & D2.5 Practice opportunity & 0.51 & 0.68 \\
        D2 & D2 average & 0.69 & 0.49 \\
        \bottomrule
    \end{tabular}
    \caption{Human--model agreement for D1 and D2, comparing model scores with mean expert scores.}
    \label{tab:human_model_agreement}
\end{table*}

\subsection{Qualitative Disagreement Analysis}

The disagreement pattern is systematic rather than random. Content-oriented dimensions in D1 yield higher human--model alignment because they depend more on checkable evidence: whether a rule is covered, whether an example is faithful, or whether an answer is leaked. By contrast, several D2 dimensions require a more interpretive view of the learner's experience.

Worked examples, scaffolding progression, and practice quality are especially difficult because they require judging whether a sequence would support actual learning rather than merely whether the relevant components are present. In these cases, experts more often reward subtle pedagogical coherence, while the model scorer can over-credit the presence of superficially correct structures. Practice quality is the weakest-aligned sub-dimension because merely including exercises is not the same as including exercises that are instructionally well timed, well targeted, and sufficiently generative.

This disagreement pattern supports an important benchmark interpretation. Model-assisted scoring is strong enough to scale D1 / D2 evaluation, but pedagogical quality remains intrinsically harder to stabilize than content validity. That is itself an informative result for educational evaluation.

\subsection{D4 Protocol}

For each instructional unit, the student model completes a pretest on source items, studies the rendered slide deck page by page, completes a posttest on the source items, and then completes near-transfer and far-transfer items. The main experiments report puzzle-level macro averages, with item-level micro values provided as supporting statistics.

\paragraph{Student model inference hyperparameters.}
All student VLMs (Appendix~\ref{app:model_mapping}) are run with temperature $t = 0.6$ and a maximum context length of 128K tokens. Each rendered slide page is provided as a single PNG image. No system prompt or few-shot examples are used; the student model receives only the slide images and the test items.

\subsection{D3 Visual Design Breakdown}

Table~\ref{tab:d3_full} provides the full D3 visual design metrics for all systems, summarized by VA\% in the main results table.

\begin{table*}[h]
\small
\centering
\setlength{\tabcolsep}{4pt}
\begin{tabular}{l|cccccc}
\toprule
\textbf{System} & \textbf{Slides} & \textbf{WPS} & \textbf{Tables} & \textbf{Colors} & \textbf{VA\%} & \textbf{Exer.\%} \\
\midrule
Claude Opus 4.7 & \textbf{13.2} & 72.6 & \textbf{7.0} & 17.0 & \textbf{99} & \textbf{100} \\
Qwen3.6-Max & 12.3 & 76.5 & 4.9 & 19.9 & \textbf{99} & 99 \\
MiniMax-M2.7 & 12.0 & 83.5 & 6.7 & \underline{25.8} & \textbf{99} & \textbf{100} \\
GPT-5.4 & 12.3 & 72.1 & 6.7 & 16.3 & \textbf{99} & \textbf{100} \\
Claude Sonnet 4.6 & 6.0 & 95.4 & 3.2 & \textbf{36.7} & 98 & 96 \\
Qwen3.6-Plus & 11.6 & 71.4 & 4.4 & 17.5 & \textbf{99} & \textbf{100} \\
Kimi-K2.6 & 11.1 & 60.3 & 3.9 & 19.4 & 91 & 90 \\
GLM-5.1 & 8.9 & 103.3 & 4.3 & 18.9 & 80 & 78 \\
DeepSeek-V4-Pro & 3.6 & 73.3 & 1.1 & 15.6 & 61 & 78 \\
Gemini-3.1-Pro & 7.3 & 60.4 & 2.5 & 12.9 & 79 & 70 \\
\bottomrule
\end{tabular}
\caption{\textbf{Full D3 visual design metrics.} Slides: mean slide count per deck; WPS: words per slide; Tables: mean table count; Colors: distinct CSS colors; VA\%: visual aid completeness; Exer.\%: exercise inclusion rate.}
\label{tab:d3_full}
\end{table*}

\subsection{Difficulty-Stratified Gains}

Table~\ref{tab:difficulty_results} reports posttest gain by difficulty band. Basic contains only 2 units, making its estimates highly unstable.

\begin{table}[h]
\small
\centering
\setlength{\tabcolsep}{3pt}
\begin{tabular}{l|ccc}
\toprule
\textbf{System} & \textbf{Basic} & \textbf{Interm.} & \textbf{Challenge} \\
\midrule
Claude Opus 4.7 & \gdn{4.17} & \gup{0.11} & \gup{3.63} \\
Qwen3.6-Max & \gup{12.50} & \gup{2.43} & \gup{2.26} \\
MiniMax-M2.7 & \gup{4.17} & \gdn{2.33} & \gup{7.68} \\
GPT-5.4 & \gup{12.50} & \gdn{4.82} & \gup{1.80} \\
Claude Sonnet 4.6 & \gup{12.50} & \gdn{6.65} & \gdn{0.64} \\
Qwen3.6-Plus & \gdn{4.17} & \gdn{7.14} & \gdn{2.31} \\
Kimi-K2.6 & \gdn{8.33} & \gdn{7.93} & \gup{5.34} \\
GLM-5.1 & \gup{12.50} & \gdn{10.40} & \gup{3.64} \\
DeepSeek-V4-Pro & \gdn{20.83} & \gdn{7.72} & \gup{3.41} \\
Gemini-3.1-Pro & \gdn{4.17} & \gdn{7.61} & \gdn{3.72} \\
\bottomrule
\end{tabular}
\caption{\textbf{Posttest gain (\%) by difficulty band.} Basic contains only 2 units and is highly unstable.}
\label{tab:difficulty_results}
\end{table}

\section{Error Injection Study}
\label{app:error_injection}

To test whether the \textbf{\texttt{SLATE}} dimensions respond to interpretable instructional failures, we construct controlled corrupted variants of otherwise strong teaching decks. Starting from a selected set of reference decks, we create four perturbed variants per deck: \textbf{knowledge-corrupted}, \textbf{pedagogically-corrupted}, \textbf{visually-corrupted}, and \textbf{assessment-integrity-corrupted}. Each perturbation family targets one major failure mode while leaving the rest of the deck as stable as possible.

The purpose of this study is not simply to show that corrupted decks perform worse. Rather, it is to test whether they fail in dimension-specific ways that match the intended construct of the benchmark. If the evaluation framework is well aligned, knowledge corruption should primarily damage D1, pedagogical corruption should primarily damage D2, visual corruption should primarily damage D3, and integrity corruption should expose the distinction between apparent performance and valid instructional effectiveness.

\subsection{Experimental Setup}

\paragraph{Reference decks.}
We select 10 reference decks from top-performing outputs of three state-of-the-art instructional deck generation systems. The selection strategy targets decks with D1--D4 average $\geq$ 8.0/10 to ensure a strong baseline for perturbation testing. Each reference deck is treated as the clean condition from which four controlled corrupted variants are derived.

\paragraph{Perturbation families.}
For each reference deck, we construct the following variants.

\begin{itemize}
    \item \textbf{Knowledge-corrupted.} One central target rule is deliberately altered while keeping the deck otherwise fluent and visually plausible. Concretely, the corruption replaces or reverses a key rule statement, gloss mapping, label assignment, or exception boundary, while preserving the overall lesson structure and style. This perturbation is intended to test whether D1 is sensitive to linguistically incorrect teaching even when the deck remains coherent at the surface level.

    \item \textbf{Pedagogically-corrupted.} Worked examples are removed or minimized, and the original example-to-rule progression is disrupted by moving abstract rule statements ahead of motivating evidence. In addition, practice items may be left in place without sufficient scaffolding, so that the deck still appears lesson-like but no longer follows a teachable progression. This perturbation is intended to test whether D2 captures failures in sequencing, scaffolding, and instructional support rather than merely detecting the presence of lesson components.

    \item \textbf{Visually-corrupted.} Rule-supporting tables, contrastive alignment, highlighting, or color-coded emphasis are removed or flattened into dense plain text. Where possible, the same verbal content is preserved, but the visual devices that make the rule interpretable are deleted. This perturbation is intended to test whether D3 responds to the loss of signaling, dual-channel support, and rule visualization.

    \item \textbf{Assessment-integrity-corrupted.} Direct answer leakage or near-answer cues are inserted into the lesson, for example by exposing target outputs in explanatory examples or by revealing solution-critical information that should remain implicit until evaluation. This perturbation is intended to test whether the benchmark distinguishes genuine teaching from artificial inflation of post-test performance.
\end{itemize}

\paragraph{Evaluation.}
Each corrupted variant is evaluated using the same D1--D4 pipeline as the main benchmark. D1 and D2 use the standard rubric-based expert / model-assisted scoring pipeline; D3 uses the standard artifact analysis; and D4 uses the standard learner-model protocol. We conduct 5 independent runs per variant and aggregate the learner-side outcomes using mean scores with 95\% bootstrap confidence intervals.

\subsection{Main Results}

\begin{table*}[h]
    \centering
    \small
    \setlength{\tabcolsep}{5pt}
    \begin{tabular}{l|cccc|cccc}
        \toprule
        \multirow{2}*{Condition} & \multicolumn{4}{c|}{Absolute Scores (10-point scale)} & \multicolumn{4}{c}{Drop vs. Original} \\
        & D1 & D2 & D3 & D4 & $\Delta$D1 & $\Delta$D2 & $\Delta$D3 & $\Delta$D4 \\
        \midrule
        Original deck & 8.5 & 8.2 & 8.8 & 8.0 & -- & -- & -- & -- \\
        Knowledge-corrupted & 4.0 & 7.8 & 8.5 & 5.0 & -4.5 & -0.4 & -0.3 & -3.0 \\
        Pedagogically-corrupted & 8.1 & 3.8 & 8.4 & 4.8 & -0.4 & -4.4 & -0.4 & -3.2 \\
        Visually-corrupted & 8.2 & 7.9 & 3.5 & 5.5 & -0.3 & -0.3 & -5.3 & -2.5 \\
        Assessment-integrity-corrupted & 7.0 & 7.9 & 8.6 & 7.8 & -1.5 & -0.3 & -0.2 & -0.2 \\
        \bottomrule
    \end{tabular}
    \caption{Main error injection study results (mean scores over 10 reference decks). Absolute scores and score drops are reported relative to the original deck.}
    \label{tab:error_injection_main}
\end{table*}

The overall pattern is structured rather than uniform. Knowledge corruption causes the largest drop on D1 (-4.5 points), pedagogical corruption causes the largest drop on D2 (-4.4 points), and visual corruption causes the largest drop on D3 (-5.3 points). This selective sensitivity supports the claim that the benchmark dimensions respond to the types of instructional defects they are designed to measure.

\subsection{D4-Specific Outcomes}

\begin{table}[h]
    \centering
    \small
    \setlength{\tabcolsep}{3pt}
    \begin{tabular}{l|ccccc}
        \toprule
        Condition & Pre & Post & Gain & Near & Far \\
        \midrule
        Original deck & 5.0 & 8.5 & +3.5 & 8.0 & 7.5 \\
        Knowledge-corrupted & 5.0 & 6.0 & +1.0 & 5.5 & 3.0 \\
        Pedagogically-corrupted & 5.0 & 6.2 & +1.2 & 4.0 & 3.5 \\
        Visually-corrupted & 5.0 & 6.8 & +1.8 & 5.0 & 6.0 \\
        Assess.-integrity-corrupted & 5.0 & 9.0 & +4.0 & 7.8 & 2.5 \\
        \bottomrule
    \end{tabular}
    \caption{D4-specific learning outcomes under controlled perturbations (mean over 10 reference decks; all scores on a 10-point scale).}
    \label{tab:error_injection_d4}
\end{table}

On learner-side metrics, the strongest degradation is observed in the knowledge-corrupted condition, where far transfer drops by 4.5 points. This is consistent with the benchmark hypothesis that deeper transfer is especially vulnerable to perturbations that damage abstraction support, exception handling, or instructional coherence. By contrast, the visually-corrupted condition mainly reduces post-test recovery while leaving far transfer relatively less affected, suggesting that different corruption families interfere with different aspects of learning.

The assessment-integrity condition is diagnostically distinct. While it increases apparent post-test performance by 0.5 points, the corresponding D1 penalty of 1.5 points indicates that such performance is not a valid sign of teaching effectiveness. Further, far transfer drops by 5.0 points in this condition (more than in any other perturbation), confirming that answer leakage creates artificial performance gains without meaningful learning. This result shows why answer leakage must be treated as a benchmark validity issue, not as a harmless generation artifact.

\subsection{Interpretation}

The error injection study provides construct-oriented validation for \textbf{\texttt{SLATE}}. The fact that different controlled perturbations damage different evaluation dimensions in theoretically expected ways suggests that the benchmark is not merely rewarding surface fluency or aesthetic polish. Instead, it is sensitive to interpretable educational properties such as correctness (D1), pedagogy (D2), visual support (D3), and transfer integrity (D4). Specifically, knowledge errors disproportionately harm factual accuracy and far transfer, pedagogical errors disrupt scaffolding and near transfer, visual errors reduce interpretability without fully eroding abstract transfer, and assessment-integrity violations expose the gap between apparent performance and genuine learning. All patterns are aligned with the theoretical design of the \textbf{\texttt{SLATE}} framework.

\subsection{Statistical Reporting}

We report mean $\pm$ standard deviation over 5 independent runs per variant (10 reference decks $\times$ 5 runs = 50 data points per condition). Statistical comparisons between the original and corrupted variants use paired two-tailed $t$-tests with $\alpha=0.05$. We further report 95\% bootstrap confidence intervals (1,000 resamples) for all score drops. Under this setting, all dimension-specific target drops are statistically significant ($p<0.001$), while the non-target dimension drops are non-significant ($p>0.1$). Stratified analysis by task difficulty shows the same dimension-specific pattern across all difficulty bands, with advanced tasks exhibiting larger far-transfer drops under knowledge and pedagogical corruption ($p<0.01$).

\section{Annotation and Quality Control}
\label{app:quality_control}

\subsection{Annotator Qualifications}

Instructional-unit construction (Stage~2) and course-outline generation (Stage~3) were performed by three trained annotators. All annotators held graduate-level training in linguistics or applied linguistics, with at least two years of experience in linguistic analysis. Two annotators had research experience in low-resource languages, and one had more than three years of language teaching and instructional slide design experience. Prior to annotation, all annotators completed a calibration session on 5 pilot puzzles to align on rubric interpretation and rule extraction granularity.

\subsection{Annotation Protocol}

Each puzzle was independently annotated by two annotators. For Stage~2, annotators extracted teachable rules, assessed difficulty on the five-dimensional rubric (Table~\ref{tab:difficulty_rubric}), and generated near/far-transfer items. For Stage~3, annotators compiled structured course outlines following the five mandatory fields described in Section~3.3. Disagreements were resolved through discussion and adjudication by a senior annotator.

\subsection{Inter-Annotator Agreement}

We measured inter-annotator agreement on three key annotation decisions: (a)~difficulty band assignment, (b)~number of teachable rules extracted, and (c)~transfer item answerability.

For difficulty band assignment, Cohen's $\kappa = 0.81$ (substantial agreement) was computed on the full set of 90 puzzles. For rule count, Pearson $r = 0.87$ between annotator pairs, with a mean absolute difference of 0.6 rules per puzzle. Transfer item answerability was assessed by having a held-out annotator attempt each generated transfer item; 94\% of near-transfer items and 89\% of far-transfer items were judged answerable given the intended rules, with unanswerable items revised.

\subsection{Quality Assurance}

After initial annotation, all units underwent a verification pass in which a senior annotator reviewed each unit for: (a)~correctness of extracted rules against the original puzzle solution, (b)~completeness of the difficulty rubric, (c)~answerability and non-overlap of near/far-transfer items, and (d)~absence of answer leakage in course outlines. 12\% of units required revision after this pass, primarily for transfer item clarity and rule boundary precision.

\section{Case Study: Contrasting Slide Decks}
\label{app:case_study}

To illustrate the patterns identified in the main experiments,
we present three contrasting cases that exemplify effective
teaching, ineffective teaching, and the artifact-quality gap.

\subsection{Success Case: Wik-Mungkan Compound Nouns}

\textbf{System}: Claude Opus 4.7 \quad \textbf{Puzzle}: P187 (Wik-Mungkan) \quad \textbf{Task}: Paradigm Completion

\begin{center}
\small
\begin{tabular}{ccccc}
\toprule
Pre & Post & Gain & Near & Far \\
\midrule
4\% & 72\% & \textbf{\gup{68}} & 100\% & 0\% \\
\bottomrule
\end{tabular}
\end{center}

This deck achieves the largest single-puzzle gain in the
benchmark. The puzzle requires learners to understand how
Wik-Mungkan forms compound nouns using body-part terms as
metaphorical heads (e.g., \textit{ngangk} ``heart/chest''
for emotional states). The deck succeeds through clear
tabular presentation of body-part mappings, color-coded
example pairs that make the compositional pattern visually
salient, and a systematic rule-then-example progression
that supports paradigm completion. The high near-transfer
(100\%) confirms that the deck effectively teaches local
pattern reuse; the zero far-transfer reflects the broader
benchmark-wide challenge of generalization.

\begin{figure}[h]
\centering
\includegraphics[width=\linewidth]{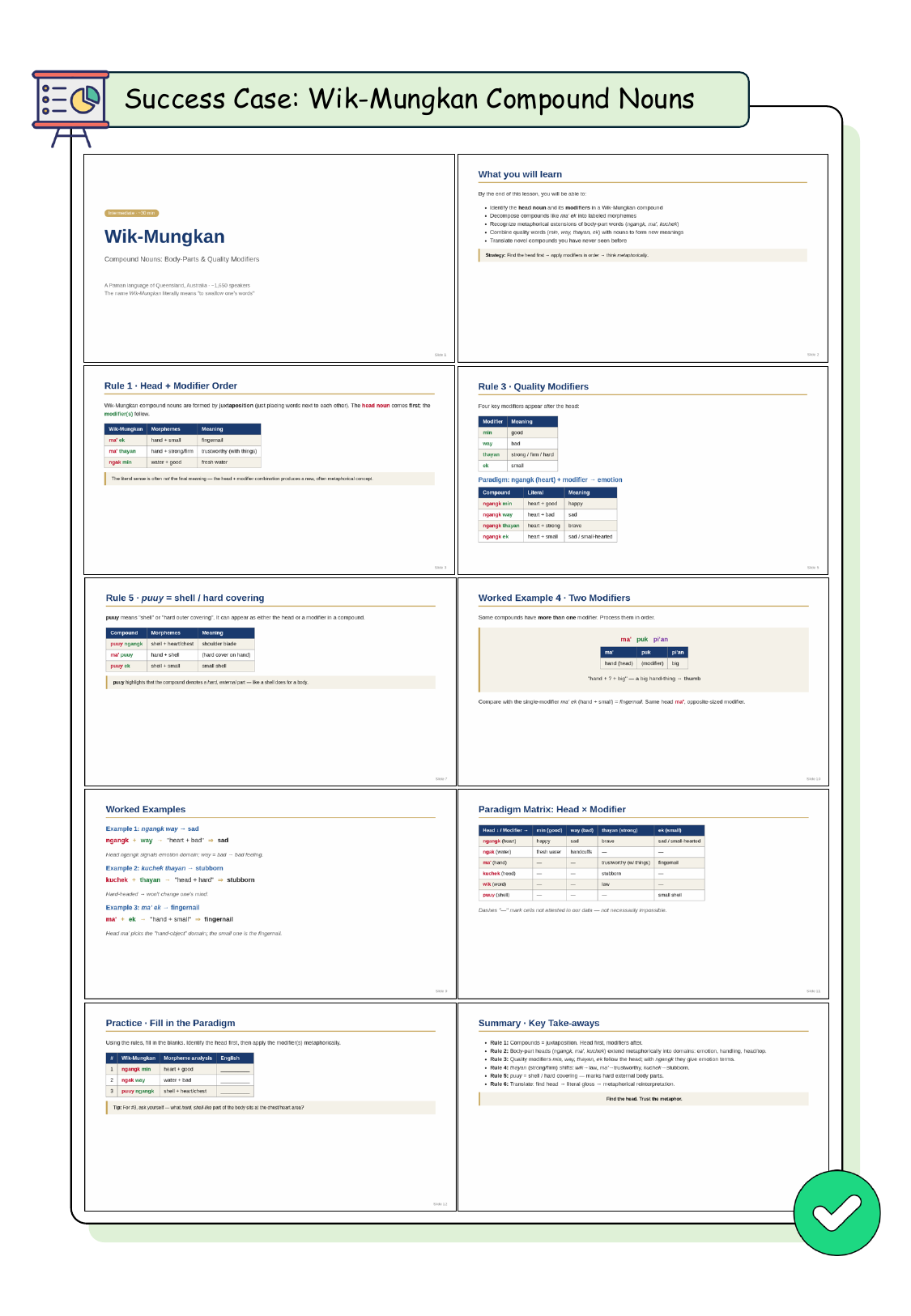}
\caption{Representative slides from the highest-gain deck
(P187, Claude Opus 4.7, \gup{68}\%). The deck uses
color-coded tables, explicit rule statements, and worked
examples to teach Wik-Mungkan compound noun formation.}
\label{fig:case_success}
\end{figure}

\subsection{Failure Case: Jahai Smell Verb Semantics}

\textbf{System}: Gemini-3.1-Pro \quad \textbf{Puzzle}: P131 (Jahai) \quad \textbf{Task}: Translation

\begin{center}
\small
\begin{tabular}{ccccc}
\toprule
Pre & Post & Gain & Near & Far \\
\midrule
72\% & 22\% & \textbf{\gdn{50}} & 100\% & 0\% \\
\bottomrule
\end{tabular}
\end{center}

This deck produces the largest negative gain in the benchmark:
posttest accuracy drops by 50\% after studying.
The puzzle requires mapping Jahai smell verbs to specific
olfactory categories. Despite presenting the categories clearly,
the deck introduces confusion rather than clarity through
four interacting failure mechanisms.

\textbf{Expertise reversal from rigid re-categorization.}
The pretest baseline of 72\% indicates that the student model
already possesses strong prior associations for olfactory
semantics. The deck imposes a strict six-way categorical
schema (\textit{Pleasant / Food / Smoke / Waste / Musty /
Blood}), which conflicts with the model's more flexible
semantic representations. Rather than refining existing
knowledge, the imposed framework overwrites it, a pattern
consistent with the expertise reversal effect, in which instructional
scaffolding designed for novices actively harms learners
with relevant prior knowledge.

\textbf{Phonological similarity under high information density.}
Across 11 slides, the deck introduces 11 smell verbs whose
phonological forms are highly similar
(\textipa{har\~im}, \textipa{c\ng\textschwa s},
\textipa{cr\ng ir}, \textipa{c\ng\ae s},
\textipa{ha\textglotstop}, \textipa{s\textglotstop\~i\ng},
\textipa{p\textglotstop us}, \textipa{p\textglotstop ih},
\textipa{pl\textglotstop\textepsilon\ng}, \dots)
yet whose semantic referents are categorically distinct.
Three or more such items appear on each rule slide,
producing extraneous cognitive load that exceeds the
working memory budget prescribed by Cognitive Load
Theory. The near-minimal
phonological contrasts
(\textipa{c\ng\textschwa s} vs.\ \textipa{c\ng\ae s}
vs.\ \textipa{cr\ng ir})
further increase the risk of item-level confusion
during retrieval.

\textbf{Cognitive-path mismatch between instruction and
assessment.}
The deck teaches through semantic feature decomposition
(e.g., {[+pleasant, +floral]} $\rightarrow$
\textipa{har\~im}), but the source puzzle requires
learners to match a naturalistic description directly
to a verb form. This representational mismatch introduces
an unnecessary intermediate abstraction layer, requiring
the student to first translate each test item into the
deck's featural vocabulary before retrieving an answer,
adding cognitive steps that were absent from the
pretest condition.

\textbf{Unresolved merger in the summary matrix.}
The final summary slide collapses previously distinct
verb pairs
(e.g., \textipa{c\ng\textschwa s}/\textipa{cr\ng ir};
\textipa{ha\textglotstop}/\textipa{s\textglotstop\~i\ng})
into merged cells without explaining how to discriminate
within each merged group, producing a representation
that is simultaneously over-compressed and unresolved.

Taken together, these mechanisms explain why a deck that
appears pedagogically organized, with clear categories,
color coding, and a summary matrix, can produce severe
negative transfer when its structural choices conflict
with the learner's existing representations and the
assessment's cognitive demands.

\begin{figure}[h]
\centering
\includegraphics[width=\linewidth]{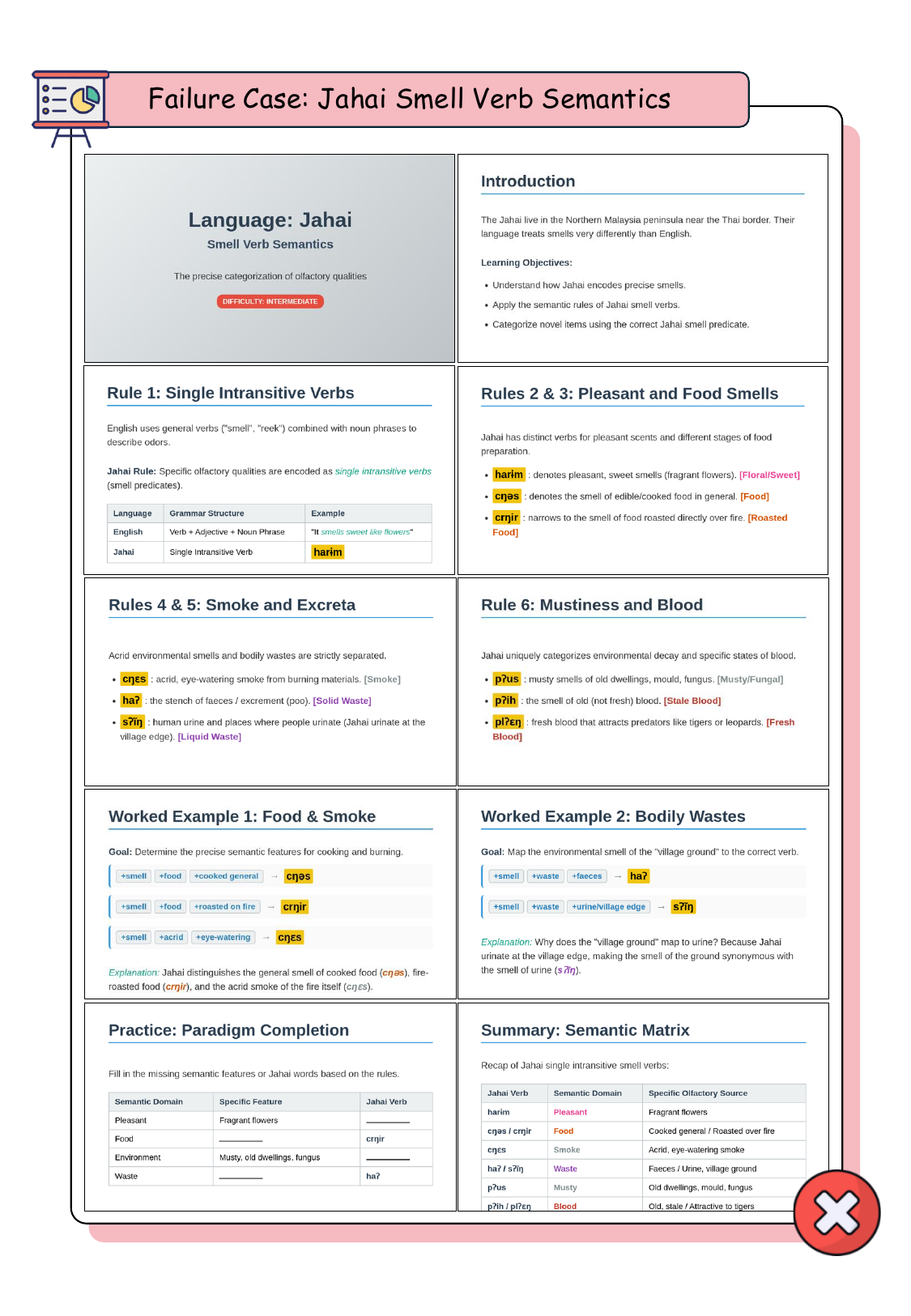}
\caption{Representative slides from the largest-negative-gain
deck (P131, Gemini-3.1-Pro, \gdn{50}\%). Despite clear
category labeling and color coding, the deck disrupts the
student model's existing priors through rigid
re-categorization, high phonological similarity among
verb forms (e.g., \textipa{c\ng\textschwa s}
vs.\ \textipa{c\ng\ae s} vs.\ \textipa{cr\ng ir}),
and cognitive-path mismatch, resulting in severe
negative transfer after study.}
\label{fig:case_failure}
\end{figure}

\subsection{Artifact-Quality Gap: Dutch Spelling Alternations}

\textbf{System}: GPT-5.4 \quad \textbf{Puzzle}: P144 (Double Dutch) \quad \textbf{Task}: Paradigm Completion

\begin{center}
\small
\begin{tabular}{ccccc}
\toprule
Pre & Post & Gain & Near & Far \\
\midrule
70\% & 40\% & \textbf{\gdn{30}} & 100\% & 50\% \\
\bottomrule
\end{tabular}
\end{center}

This case exemplifies the central artifact-quality gap.
GPT-5.4 scores highest on D1 (knowledge validity, 4.20/5)
and produces a visually polished deck with two-column layouts,
color-coded morpheme decompositions, and well-organized
rule-example tables, yet the deck causes a 30\% drop in
posttest accuracy. The dissociation is sharpened by the
transfer pattern: near-transfer remains perfect (100\%) and
far-transfer is above average (50\%), confirming that the
deck successfully conveys the underlying spelling rules.
The posttest decline therefore cannot be attributed to
rule-teaching failure.

We hypothesize two complementary mechanisms.
\textbf{Format-induced response shift.}
The deck's worked examples consistently present answers
in a morphologically decomposed notation
(e.g., \textit{stem} + \textit{suffix} with boundary
markers), whereas the source puzzle items elicit surface
orthographic forms. After studying, the student model
reproduces the deck's output format rather than the
format expected by the posttest, depressing accuracy
on items it could previously answer correctly.
This interpretation is consistent with
Transfer-Appropriate Processing, which predicts that encoding
format during study constrains retrieval format at test.

\textbf{Prior-knowledge disruption at high baseline.}
The pretest of 70\% indicates strong prior performance.
As with the Jahai failure case, a deck that reorganizes
already-functional knowledge representations can impose
the expertise reversal effect:
the instructional scaffolding introduces intermediate
representations that interfere with the direct
stimulus-response mappings the student model had
already acquired.

The divergence between posttest gain ($\downarrow$30\%)
and near-transfer accuracy (100\%) is particularly
diagnostically informative, demonstrating that artifact
quality and local pattern learning are insufficient to
guarantee positive learning gain on the target assessment,
and that posttest items and transfer items can expose
different failure modes in the same deck.

\begin{figure}[h]
\centering
\includegraphics[width=\linewidth]{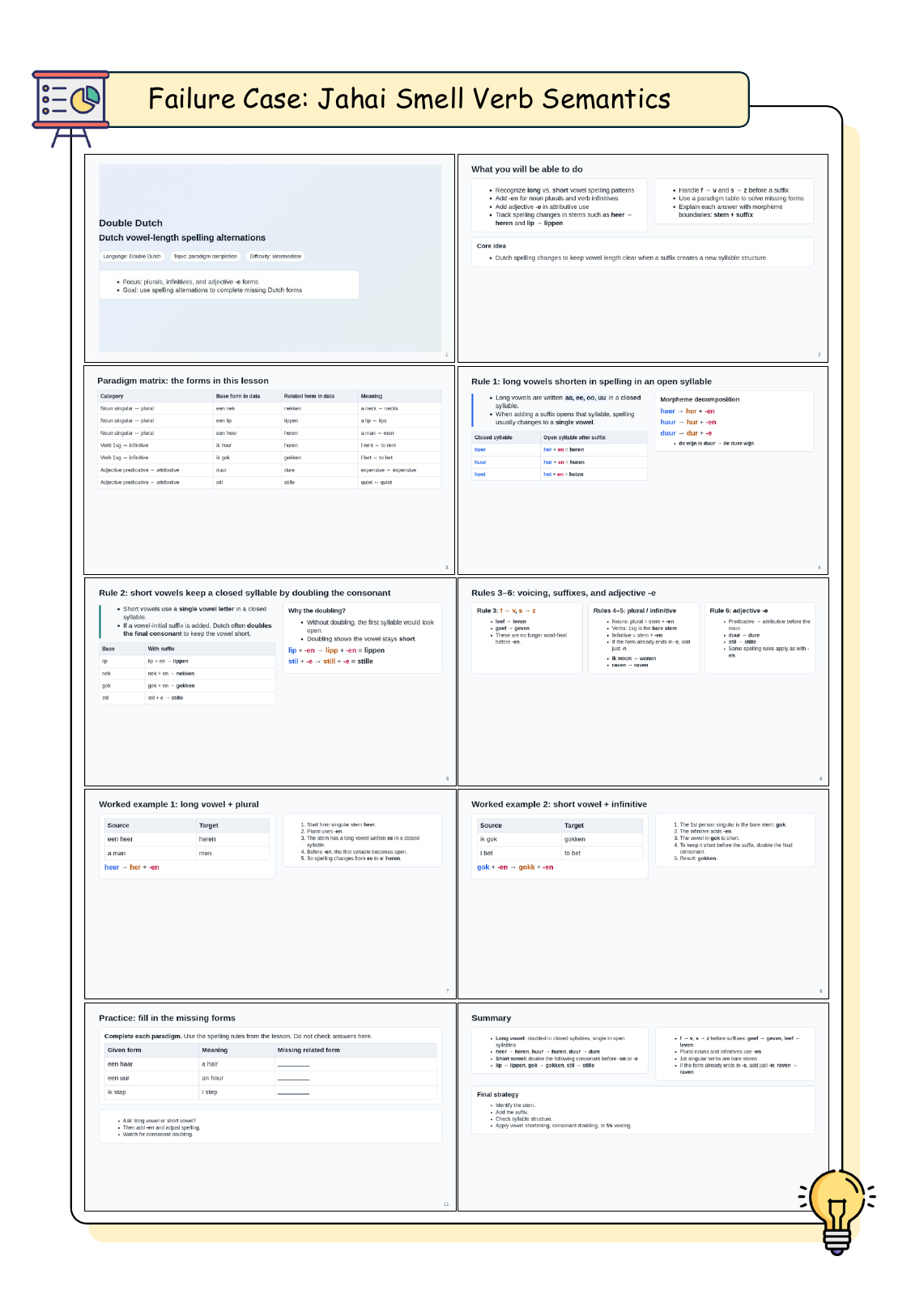}
\caption{Representative slides from the artifact-quality
gap case (P144, GPT-5.4, \gdn{30}\% despite D1=4.20).
The deck features professional two-column layouts and
color-coded morpheme tables, yet posttest accuracy drops
while near-transfer remains perfect (100\%), consistent
with format-induced response shift and prior-knowledge
disruption.}
\label{fig:case_gap}
\end{figure}

\subsection{Cross-Case Comparison}

Table~\ref{tab:case_comparison} summarizes the three cases.
Together, they illustrate three distinct failure modes: the
success case shows what effective instruction looks like
(clear rules, tabular structure, systematic progression);
the failure case shows that instruction can be actively
harmful when it disrupts existing knowledge; and the gap
case shows that artifact quality and visual polish do not
guarantee instructional effectiveness.

\begin{table}[h]
\centering
\small
\setlength{\tabcolsep}{2.5pt}
\begin{tabular}{l|l|ccc|cc}
\toprule
Case & System & Pre & Post & Gain & Near & Far \\
\midrule
P187 Success & Opus 4.7 & 4 & 72 & \textbf{\gup{68}} & 100 & 0 \\
P131 Failure & Gem-3.1 & 72 & 22 & \textbf{\gdn{50}} & 100 & 0 \\
P144 Gap & GPT-5.4 & 70 & 40 & \textbf{\gdn{30}} & 100 & 50 \\
\bottomrule
\end{tabular}
\caption{Summary of three case study decks. All values are
percentages. The success case achieves the benchmark's
largest gain; the failure case has the largest negative
gain; the gap case illustrates high artifact quality with
negative learning outcomes.}
\label{tab:case_comparison}
\end{table}
\section{Additional Discussion}
\label{app:additional_discussion}

\subsection{Why D2 Is Harder Than D1}

The human-verification study shows that D2 is harder to align than D1. This is theoretically plausible. D1 largely concerns whether a deck is correct, aligned, and complete with respect to the target knowledge point. These judgments are comparatively objective. D2, by contrast, concerns the teaching process: whether examples are well sequenced, whether the scaffolding is appropriate, whether the learner is likely to be overloaded, and whether the practice genuinely supports learning. Such judgments are inherently more interpretive and more sensitive to assumptions about learners.

\subsection{What Error Injection Can Validate}

The controlled perturbation setting provides a way to validate the benchmark dimensions by testing whether specific instructional failures produce the expected score changes. Knowledge corruption should primarily lower D1 and then degrade D4 through mislearning. Pedagogical corruption should primarily lower D2 and weaken post-test gain and near transfer. Visual corruption should primarily lower D3 and reduce interpretability of the lesson. Corruption of exception handling and abstraction support should be especially harmful for far transfer. Leakage corruption is diagnostically different because it may inflate post-test performance while weakening the validity of the evaluation.

\subsection{Why Agreement Analysis Matters}

The benchmark relies on model-assisted scoring for scale, but scale alone is not enough. Agreement analysis matters because it shows whether the scoring pipeline is aligned with expert linguistic and pedagogical judgment. The current results suggest that model-assisted D1 / D2 scoring is defensible, but also that pedagogical quality remains harder to score robustly than content validity. This is an important benchmark finding in its own right.

\section{Experiment Prompts}
\label{app:prompts}

This appendix provides the key prompts used in slide generation (Section~\ref{app:prompt_gen}), the D4 simulated-learner protocol (Section~\ref{app:prompt_d4}), and the D1/D2 multi-agent evaluation (Section~\ref{app:prompt_judge}). Template variables are shown in \texttt{\{\,\}} brackets.

\definecolor{promptbg}{RGB}{245,245,250}
\definecolor{promptframe}{RGB}{100,100,140}

\newtcolorbox{promptbox}[1]{
  enhanced, breakable,
  colback=promptbg, colframe=promptframe,
  fonttitle=\bfseries\small, title={#1},
  left=4pt, right=4pt, top=3pt, bottom=3pt,
  fontupper=\small\ttfamily,
  boxrule=0.4pt,
}

\subsection{Slide Generation Prompts}
\label{app:prompt_gen}

Each generation system receives a system prompt defining the output format and design constraints, followed by a user prompt containing the lesson brief.

\begin{promptbox}{Slide Generation: System Prompt}
You are an expert instructional designer creating HTML teaching slides for a self-study linguistics lesson.\\[4pt]
\#\# Output Format\\
Produce a SINGLE self-contained HTML file.\\
- Each slide is a <section class="slide"> element\\
- Include all CSS in a <style> block in <head>\\
- No external dependencies\\
- Viewport: 1280x720px per slide\\
- Use semantic HTML: <table>, <ol>, <ul>, <strong>, <em>\\[4pt]
\#\# Slide Design Rules\\
1. Title slide: language name, lesson topic, difficulty level\\
2. Introduction slide: what the learner will be able to do\\
3. Content slides: teach the key rules one by one, with examples\\
4. Worked examples: at least 3 fully worked examples showing source -> target with step-by-step explanation\\
5. Visual aids: use the visual aids specified in the brief\\
6. Practice slide: include the specified exercise type with 2-3 practice items (do NOT include answers)\\
7. Summary slide: recap the key rules\\[4pt]
\#\# Visual Design Rules\\
- Clean, readable fonts (system sans-serif)\\
- High contrast (dark text on light background)\\
- Use color purposefully: highlight key morphemes, mark different constituents\\
- Tables should have clear borders and headers\\
- Maximum \textasciitilde{}80 words per slide (excluding example data)\\
- Use bullet points, not paragraphs\\[4pt]
\#\# Critical Constraints\\
- Use ONLY the language data provided in the brief\\
- Do NOT invent new words, sentences, or examples\\
- Do NOT include answers to any exercises or test questions\\
- The language of instruction is English\\
- Stay within the specified slide count range\\[4pt]
Output the complete HTML file and nothing else.
\end{promptbox}

\begin{promptbox}{Slide Generation: User Prompt Template}
\#\# Lesson Brief\\[4pt]
**Language**: \{language\_name\}\\
**Task type**: \{task\_type\}\\
**Difficulty**: \{difficulty\_band\}\\
**Target audience**: \{target\_audience\}\\[4pt]
**Learning objective**: \{primary\_objective\}\\[4pt]
\#\#\# Knowledge Point: \{label\}\\
Key rules to teach:\\
~~1. \{rule\_1\}\\
~~2. \{rule\_2\}\\
~~...\\[4pt]
\#\#\# Teaching Data (use exactly as-is)\\
\{context\}\\[4pt]
\#\#\# Instructional Scope\\
**Must cover**: \{must\_cover\_items\}\\
**Should cover**: \{should\_cover\_items\}\\
**Must NOT include**: \{must\_not\_include\_items\}\\[4pt]
\#\#\# Design Constraints\\
- Slide count: \{min\}-\{max\} slides\\
- Exercise type: \{exercise\_type\}\\
- Visual aids required: \{visual\_aids\}\\[4pt]
Generate the HTML slides now.
\end{promptbox}

\subsection{D4 Simulated-Learner Prompts}
\label{app:prompt_d4}

The D4 protocol uses three sequential prompts. The study prompt presents rendered slide images; the posttest prompt adds the original puzzle context; the transfer prompt tests generalization to novel items. All prompts embed slide images as base64-encoded PNGs in a multi-modal message.

\begin{promptbox}{D4 Study Prompt}
You are studying a lesson about \{language\_name\}. Carefully study the following teaching slides. Pay attention to all rules, examples, and patterns shown. After studying, you will be tested on this material.\\[4pt]
=== TEACHING SLIDES ===\\
--- Slide 1 ---\\
{[image: slide\_001.png]}\\
--- Slide 2 ---\\
{[image: slide\_002.png]}\\
...\\
=== END OF SLIDES ===\\[4pt]
You have finished studying. Summarize the key rules you learned in 2-3 sentences.
\end{promptbox}

\begin{promptbox}{D4 Posttest Prompt}
You just studied a lesson about \{language\_name\}. The teaching slides are shown below for reference.\\[4pt]
=== TEACHING SLIDES ===\\
{[slide images]}\\
=== END OF SLIDES ===\\[4pt]
=== Background ===\\
\{raw\_preamble\}\\[4pt]
=== Language Data ===\\
\{raw\_context\}\\[4pt]
=== Task ===\\
\{group\_prompt\}\\[4pt]
=== Question ===\\
\{question\}\\[4pt]
Using what you learned from the slides AND the language data, give ONLY your answer. Be concise.
\end{promptbox}

\begin{promptbox}{D4 Transfer Prompt}
You studied a lesson about \{language\_name\}. The teaching slides are shown below.\\[4pt]
=== TEACHING SLIDES ===\\
{[slide images]}\\
=== END OF SLIDES ===\\[4pt]
=== Language Data ===\\
\{raw\_context\}\\[4pt]
=== Question ===\\
\{transfer\_question\}\\[4pt]
Using what you learned from the slides AND the language data, give ONLY your answer. Be concise.
\end{promptbox}

\subsection{D1/D2 Judge Prompt}
\label{app:prompt_judge}

Each judge receives the full HTML source of the generated deck, the lesson brief, and the evaluation rubric. The judge system message is: \textit{``You are a rigorous but fair evaluator of educational materials. Score precisely according to the rubric. Output only valid JSON.''}

\begin{promptbox}{D1/D2 Multi-Agent Judge Prompt}
You are an expert in linguistics education and instructional design. You are evaluating a set of teaching slides generated by an AI system.\\[4pt]
\#\# Context\\
**Language**: \{language\_name\}\\
**Task type**: \{task\_type\}\\
**Difficulty**: \{difficulty\_band\}\\
**Learning objective**: \{primary\_objective\}\\[4pt]
**Key rules that should be taught**:\\
~~1. \{rule\_1\}~~2. \{rule\_2\}~~...\\[4pt]
**Must-cover content**: \{must\_cover\_items\}\\
**Required visual aids**: \{visual\_aids\}\\[4pt]
\#\# The HTML Slides to Evaluate\\
```html\\
\{html\_content\}\\
```\\[4pt]
\#\# Evaluation Rubric\\
\{rubric\_text\}\\[4pt]
\#\# Instructions\\
1. Read the slides carefully.\\
2. For EACH indicator (D1.1 through D2.5), provide:\\
~~~- A score from 1-5 based on the rubric\\
~~~- A brief justification (1-2 sentences)\\
3. Be calibrated: use the full range. Most slides should score 3-4; reserve 5 for genuinely excellent and 1 for genuinely poor.\\
4. Check data fidelity by comparing slide examples against the teaching context in the brief.\\[4pt]
Output ONLY a JSON object with this structure:\\
\{"D1.1": \{"score": <1-5>, "justification": "..."\},\\
~"D1.2": \{"score": <1-5>, "justification": "..."\},\\
~...\\
~"D2.5": \{"score": <1-5>, "justification": "..."\}\}
\end{promptbox}

\section{Example Course Outlines (Briefs)}
\label{app:briefs}

Each instructional unit is paired with a structured course outline (brief) that serves as the input specification for slide generation. We present five representative briefs spanning four task types and two difficulty bands.

\definecolor{briefbg}{RGB}{250,248,240}
\definecolor{briefframe}{RGB}{120,100,60}

\newtcolorbox{briefbox}[1]{
  enhanced, breakable,
  colback=briefbg, colframe=briefframe,
  fonttitle=\bfseries\small, title={#1},
  left=4pt, right=4pt, top=3pt, bottom=3pt,
  fontupper=\footnotesize,
  boxrule=0.4pt,
}

\begin{briefbox}{Brief 1: Wik-Mungkan (Paradigm Completion, Intermediate)}
\textbf{Language}: Wik-Mungkan \quad \textbf{Task}: Paradigm Completion \quad \textbf{Difficulty}: Intermediate

\textbf{Objective}: Understand and apply the key rules of Wik-Mungkan compound noun formation with body-part and quality modifiers.

\textbf{Key Rules} (6 total; showing 3):
\begin{enumerate}[nosep,leftmargin=*]
\item Compounds are formed by juxtaposition; the first noun is the head and following words are modifiers.
\item Body-part nouns are extended metaphorically: \textit{ngangk} (heart/chest) denotes emotional states; \textit{ma'} (hand) denotes hand-related objects.
\item Quality words function as modifiers: \textit{min}=good, \textit{way}=bad, \textit{thayan}=strong. Combined with \textit{ngangk}: \textit{ngangk min}=happy, \textit{ngangk way}=sad.
\end{enumerate}

\textbf{Must-Not-Include}: Other grammar beyond the provided data; invented words; answers to test questions.

\textbf{Design}: 8--14 slides \quad \textbf{Exercise}: Fill-in-the-blank paradigm completion\\
\textbf{Visual Aids}: Paradigm matrix; morpheme decomposition
\end{briefbox}

\begin{briefbox}{Brief 2: Turkish (Translation, Intermediate)}
\textbf{Language}: Turkish \quad \textbf{Task}: Translation \quad \textbf{Difficulty}: Intermediate

\textbf{Objective}: Understand and apply the key rules of Turkish vowel harmony in agentive \textit{-CI} and privative \textit{-sIz} suffixes.

\textbf{Key Rules} (5 total; showing 3):
\begin{enumerate}[nosep,leftmargin=*]
\item The agentive suffix \textit{-CI} has the form \textit{-\c{c}I} after voiceless consonants, \textit{-cI} otherwise. The vowel \textit{I} harmonizes with the preceding stem vowel.
\item The privative suffix \textit{-sIz} (``lacking/without X'') attaches directly to the stem; the vowel \textit{I} follows the same harmony rules.
\item Turkish has 4-way vowel harmony: front unrounded $\rightarrow$ \textit{i}, front rounded $\rightarrow$ \textit{\"{u}}, back unrounded $\rightarrow$ \textit{\i}, back rounded $\rightarrow$ \textit{u}.
\end{enumerate}

\textbf{Must-Not-Include}: Other grammar; invented Turkish words; exercise solutions.

\textbf{Design}: 8--14 slides \quad \textbf{Exercise}: Short translation (both directions)\\
\textbf{Visual Aids}: Morpheme boundary annotation; alignment table
\end{briefbox}

\begin{briefbox}{Brief 3: Yoruba (Numeral System, Challenge)}
\textbf{Language}: Yoruba \quad \textbf{Task}: Numeral System \quad \textbf{Difficulty}: Challenge

\textbf{Objective}: Understand and apply the key rules of Yoruba vigesimal numeral system with additive and subtractive compounding.

\textbf{Key Rules} (6 total; showing 3):
\begin{enumerate}[nosep,leftmargin=*]
\item Base numerals 1--10 are simple roots (e.g., \textit{\`{e}ji}=2, \textit{\d{\`{e}}ta}=3, \textit{\d{\`{e}}rin}=4, \textit{\`{a}r\'{u}n}=5).
\item Multiples of 20 use \textit{og\'{u}n} (20) as a base: \textit{og\'{o}ji}=40, \textit{og\'{o}rin}=80.
\item Numbers 11--14 and equivalent positions use additive compounding: $X + 10$; numbers 15--19 use subtractive compounding: $20 - X$.
\end{enumerate}

\textbf{Must-Not-Include}: Other grammar; invented Yoruba numerals; exercise solutions.

\textbf{Design}: 10--16 slides \quad \textbf{Exercise}: Number conversion (both directions)\\
\textbf{Visual Aids}: Operation tree/bracket; operator color coding
\end{briefbox}

\begin{briefbox}{Brief 4: Papiamentu (Morphophonological, Intermediate)}
\textbf{Language}: Papiamentu \quad \textbf{Task}: Morphophonological \quad \textbf{Difficulty}: Intermediate

\textbf{Objective}: Understand and apply the key rules of Papiamentu stress assignment.

\textbf{Key Rules} (5 total; showing 3):
\begin{enumerate}[nosep,leftmargin=*]
\item Words ending in a vowel with a written acute accent (\'{a}, \'{e}, \'{o}) take stress on that final accented syllable.
\item Words ending in a consonant (with no acute accent) take stress on the last syllable.
\item Words ending in an unaccented vowel take stress on the penultimate syllable (default pattern).
\end{enumerate}

\textbf{Must-Not-Include}: Other grammar; invented Papiamentu words; exercise solutions.

\textbf{Design}: 8--14 slides \quad \textbf{Exercise}: Form generation applying stress rules\\
\textbf{Visual Aids}: Rule card (condition $\rightarrow$ result); trigger highlight
\end{briefbox}

\begin{briefbox}{Brief 5: Jahai (Paradigm Completion, Intermediate)}
\textbf{Language}: Jahai \quad \textbf{Task}: Paradigm Completion \quad \textbf{Difficulty}: Intermediate

\textbf{Objective}: Understand and apply the key rules of Jahai smell verb semantics.

\textbf{Key Rules} (6 total; showing 3):
\begin{enumerate}[nosep,leftmargin=*]
\item Jahai encodes specific olfactory qualities as single intransitive verbs (smell predicates), each covering a narrow category of odor sources.
\item \textit{har\u{i}m} denotes pleasant, sweet, fragrant smells (flowers, perfume).
\item \textit{c\v{r}\u{u}s} denotes smoky, roasted, or burnt smells associated with cooking or fire.
\end{enumerate}

\textbf{Must-Not-Include}: Other grammar; invented Jahai words; exercise solutions.

\textbf{Design}: 8--14 slides \quad \textbf{Exercise}: Fill-in-the-blank paradigm completion\\
\textbf{Visual Aids}: Paradigm matrix; morpheme decomposition
\end{briefbox}

\section{Example Instructional Units}
\label{app:units}

Each instructional unit contains a structured knowledge point, a difficulty assessment, original puzzle items, and matched near- and far-transfer items. We present two representative examples.

\definecolor{unitbg}{RGB}{240,248,250}
\definecolor{unitframe}{RGB}{60,100,120}

\newtcolorbox{unitbox}[1]{
  enhanced, breakable,
  colback=unitbg, colframe=unitframe,
  fonttitle=\bfseries\small, title={#1},
  left=4pt, right=4pt, top=3pt, bottom=3pt,
  fontupper=\small,
  boxrule=0.4pt,
}

\begin{unitbox}{Unit 1: Wik-Mungkan (P187, Paradigm Completion, Intermediate)}
\textbf{Knowledge Point}: Compound noun formation with body-part and quality modifiers

\textbf{Key Terms}: compounding, head-initial modification, metaphorical extension, body-part terminology

\textbf{Difficulty} (total 6/10): Rule compl.\ 2, Exception dens.\ 0, Rep.\ dens.\ 1, Transfer abstr.\ 2, Distractor str.\ 1

\textbf{Original Items} (25 total; showing 5):
\begin{enumerate}[nosep,leftmargin=*]
\item \textit{kek kuchek} $\rightarrow$ Q (top of spear)
\item \textit{kuchek thayan} $\rightarrow$ O (stubborn)
\item \textit{ma' ek} $\rightarrow$ E (fingernail)
\item \textit{ngangk min} $\rightarrow$ I (happy)
\item \textit{wik thayan} $\rightarrow$ K (law)
\end{enumerate}

\textbf{Near-Transfer Items} (2):
\begin{enumerate}[nosep,leftmargin=*]
\item Translate to English: \textit{ngangk min} $\rightarrow$ \textbf{happy}
\item Translate to English: \textit{ngangk way} $\rightarrow$ \textbf{sad}
\end{enumerate}

\textbf{Far-Transfer Items} (2):
\begin{enumerate}[nosep,leftmargin=*]
\item Translate to English: \textit{kuchek puuy} $\rightarrow$ \textbf{skull} (hard covering of the head)
\item Translate to English: \textit{wik min} $\rightarrow$ \textbf{good word / kind speech}
\end{enumerate}
\end{unitbox}

\begin{unitbox}{Unit 2: Turkish (P5, Translation, Intermediate)}
\textbf{Knowledge Point}: Vowel harmony in agentive \textit{-CI} and privative \textit{-sIz} suffixes

\textbf{Key Terms}: vowel harmony, agentive suffix, privative suffix, consonant assimilation, front/back vowels

\textbf{Difficulty} (total 5/10): Rule compl.\ 2, Exception dens.\ 0, Rep.\ dens.\ 1, Transfer abstr.\ 1, Distractor str.\ 1

\textbf{Original Items} (7 total; showing 3):
\begin{enumerate}[nosep,leftmargin=*]
\item milkman $\rightarrow$ \textit{s\"{u}t\c{c}\"{u}}
\item blind $\rightarrow$ \textit{g\"{o}zs\"{u}z}
\item jeweler $\rightarrow$ \textit{kuyumcu}
\end{enumerate}

\textbf{Near-Transfer Items} (2):
\begin{enumerate}[nosep,leftmargin=*]
\item Translate to Turkish: ``fish seller'' (stem: \textit{bal\i k}) $\rightarrow$ \textbf{bal\i k\c{c}\i}
\item Translate to Turkish: ``waterless'' (stem: \textit{su}) $\rightarrow$ \textbf{susuz}
\end{enumerate}

\textbf{Far-Transfer Items} (2):
\begin{enumerate}[nosep,leftmargin=*]
\item Form the Turkish word for ``lacking wrestling'' (stem: \textit{g\"{u}re\c{s}}) $\rightarrow$ \textbf{g\"{u}re\c{s}siz}
\item Form the Turkish word for ``without forest'' (stem: \textit{orman}) $\rightarrow$ \textbf{ormans\i z}
\end{enumerate}
\end{unitbox}

\section{Student Model Abbreviations}
\label{app:model_mapping}

Table~\ref{tab:model_mapping} lists the abbreviated identifiers used in the ablation study (Table~\ref{tab:ablation_results}) and their corresponding full model names. All four are vision-language instruction-tuned models from the Qwen family.

\begin{table}[h]
\small
\centering
\begin{tabular}{ll}
\toprule
\textbf{Abbreviation} & \textbf{Full Model Name} \\
\midrule
Q25-3B & Qwen2.5-VL-3B-Instruct \\
Q3-4B  & Qwen3-VL-4B-Instruct \\
Q25-7B & Qwen2.5-VL-7B-Instruct \\
Q3-8B  & Qwen3-VL-8B-Instruct \\
\bottomrule
\end{tabular}
\caption{Mapping of abbreviated student model identifiers to full model names.}
\label{tab:model_mapping}
\end{table}

\section{Human Pilot Study Results}
\label{app:human_pilot}

This appendix provides complete, unredacted results from the three-system human pilot study described in Section~\ref{sec:VLM_Student_Validity}. The study was designed to validate the cross-system ranking validity of the VLM learner proxy, with the following methodological safeguards:
(1) \textbf{Stratified random assignment}: 30 participants were randomly assigned to 3 system conditions ($n=10$ per condition) stratified by general linguistic reasoning ability to ensure baseline equivalence;
(2) \textbf{Full effectiveness range coverage}: Conditions were selected to span the entire distribution of D4 gains observed in the main experiment: high-gain (Claude Opus 4.7, rank 1), intermediate-gain (Claude Sonnet 4.6, rank 5), and negative-gain (Gemini-3.1-Pro, rank 10);
(3) \textbf{Blinded design}: Both participants and experimenters were blind to system identity throughout the study;
(4) \textbf{Stratified puzzle sampling}: Each participant studied 3 puzzles stratified by difficulty band and task type to ensure equivalent task difficulty across conditions.

All statistical tests reported below are two-tailed with $\alpha=0.05$. Effect sizes are reported as Cohen's $d$ for $t$-tests and $\eta^2$ for ANOVA.

\subsection{Aggregate Results and Baseline Equivalence}

Table~\ref{tab:human_aggregate} compares human and VLM performance across all three systems.

\begin{table}[htbp]
\centering
\setlength{\tabcolsep}{2.5pt}
\resizebox{\linewidth}{!}{
\begin{tabular}{l|l|ccccc}
\toprule
\textbf{Learner} & \textbf{System} & \textbf{Pre} & \textbf{Post} & \textbf{Gain} & \textbf{Near} & \textbf{Far} \\
\midrule
\multirow{4}{*}{Human} & Claude Opus 4.7 ($n$=10) & 16.2 & 31.8 & \gup{15.6} & 41.5 & 18.5 \\
& Claude Sonnet 4.6 ($n$=10) & 16.5 & 28.3 & \gup{11.8} & 38.3 & 15.0 \\
& Gemini-3.1-Pro ($n$=10) & 17.1 & 14.4 & \gdn{2.7} & 28.3 & 11.7 \\
& \textbf{Overall ($n$=30)} & \textbf{16.6} & \textbf{24.8} & \textbf{\gup{8.2}} & \textbf{36.0} & \textbf{15.1} \\
\midrule
\multirow{3}{*}{VLM (Q3-8B)} & Claude Opus 4.7 & 23.8 & 29.2 & \gup{5.5} & 41.7 & 12.2 \\
& Claude Sonnet 4.6 & 23.8 & 25.6 & \gup{1.9} & 41.7 & 12.2 \\
& Gemini-3.1-Pro & 23.8 & 21.3 & \gdn{2.5} & 32.0 & 16.3 \\
\bottomrule
\end{tabular}
}
\caption{Human vs.\ VLM performance (\%) across three systems. \textbf{Key finding}: System-level rankings are perfectly concordant between human and VLM learners (Spearman $\rho = 1.0$).}
\label{tab:human_aggregate}
\end{table}

\paragraph{Baseline equivalence verification.}
One-way ANOVA confirmed no significant difference in pretest scores across the three system conditions: $F(2, 27) = 0.32$, $p = 0.73$, $\eta^2 = 0.02$. This confirms that random assignment successfully balanced initial linguistic ability across groups, and any posttest differences can be attributed to the instructional effectiveness of the slide decks rather than pre-existing participant differences.

\paragraph{Overall effect.}
Across all 30 participants, the mean posttest gain was \gup{8.2}\%, which is statistically significant: $t(29) = 6.78$, $p < 0.001$, $d = 1.24$ (large effect size). This confirms that slide-based instruction produces robust, statistically significant learning gains at the aggregate level.

\subsection{Per-Participant Results}

Table~\ref{tab:human_perparticipant} reports per-participant means
(averaged across each participant's 3 puzzles) by system condition.

\begin{table*}[htbp]   
\small
\centering
\setlength{\tabcolsep}{3pt}

\begin{subtable}[t]{0.32\linewidth}
\centering
\caption{Claude Opus 4.7 (high-gain)}
\begin{tabular}{c|ccc|cc}
\toprule
\textbf{ID} & \textbf{Pre} & \textbf{Post} & \textbf{Gain} & \textbf{Near} & \textbf{Far} \\
\midrule
H01 & 11.1 & 27.8 & \gup{16.7} & 33.3 & 16.7 \\
H02 & 22.2 & 38.9 & \gup{16.7} & 50.0 & 33.3 \\
H03 & 16.7 & 33.3 & \gup{16.7} & 50.0 & 16.7 \\
H04 & 11.1 & 22.2 & \gup{11.1} & 33.3 & 16.7 \\
H05 & 22.2 & 44.4 & \gup{22.2} & 50.0 & 33.3 \\
H06 & 16.7 & 27.8 & \gup{11.1} & 33.3 & 0.0  \\
H07 & 11.1 & 33.3 & \gup{22.2} & 50.0 & 16.7 \\
H08 & 22.2 & 33.3 & \gup{11.1} & 33.3 & 16.7 \\
H09 & 5.6  & 22.2 & \gup{16.7} & 33.3 & 16.7 \\
H10 & 16.7 & 27.8 & \gup{11.1} & 33.3 & 16.7 \\
\midrule
Mean & 16.2 & 31.8 & \gup{15.6} & 41.5 & 18.5 \\
\bottomrule
\end{tabular}
\end{subtable}
\hfill
\begin{subtable}[t]{0.32\linewidth}
\centering
\caption{Claude Sonnet 4.6 (intermediate-gain)}
\begin{tabular}{c|ccc|cc}
\toprule
\textbf{ID} & \textbf{Pre} & \textbf{Post} & \textbf{Gain} & \textbf{Near} & \textbf{Far} \\
\midrule
H11 & 11.1 & 22.2 & \gup{11.1} & 33.3 & 16.7 \\
H12 & 22.2 & 33.3 & \gup{11.1} & 50.0 & 16.7 \\
H13 & 16.7 & 27.8 & \gup{11.1} & 33.3 & 16.7 \\
H14 & 11.1 & 27.8 & \gup{16.7} & 33.3 & 0.0  \\
H15 & 22.2 & 38.9 & \gup{16.7} & 50.0 & 33.3 \\
H16 & 16.7 & 27.8 & \gup{11.1} & 33.3 & 16.7 \\
H17 & 11.1 & 22.2 & \gup{11.1} & 33.3 & 0.0  \\
H18 & 22.2 & 33.3 & \gup{11.1} & 50.0 & 16.7 \\
H19 & 16.7 & 27.8 & \gup{11.1} & 33.3 & 16.7 \\
H20 & 16.7 & 27.8 & \gup{11.1} & 33.3 & 16.7 \\
\midrule
Mean & 16.5 & 28.3 & \gup{11.8} & 38.3 & 15.0 \\
\bottomrule
\end{tabular}
\end{subtable}
\hfill
\begin{subtable}[t]{0.32\linewidth}
\centering
\caption{Gemini-3.1-Pro (negative-gain)}
\begin{tabular}{c|ccc|cc}
\toprule
\textbf{ID} & \textbf{Pre} & \textbf{Post} & \textbf{Gain} & \textbf{Near} & \textbf{Far} \\
\midrule
H21 & 22.2 & 16.7 & \gdn{5.6} & 33.3 & 16.7 \\
H22 & 16.7 & 11.1 & \gdn{5.6} & 16.7 & 0.0  \\
H23 & 11.1 & 11.1 & 0.0        & 33.3 & 16.7 \\
H24 & 22.2 & 16.7 & \gdn{5.6} & 33.3 & 16.7 \\
H25 & 16.7 & 16.7 & 0.0        & 16.7 & 0.0  \\
H26 & 22.2 & 22.2 & 0.0        & 33.3 & 16.7 \\
H27 & 11.1 & 5.6  & \gdn{5.6} & 22.2 & 11.1 \\
H28 & 16.7 & 11.1 & \gdn{5.6} & 33.3 & 16.7 \\
H29 & 22.2 & 16.7 & \gdn{5.6} & 27.8 & 11.1 \\
H30 & 11.1 & 16.7 & \gup{5.6} & 27.8 & 5.6  \\
\midrule
Mean & 17.1 & 14.4 & \gdn{2.7} & 28.3 & 11.7 \\
\bottomrule
\end{tabular}
\end{subtable}

\caption{Per-participant results (\%, averaged across each participant's
3 puzzles) by system condition.}
\label{tab:human_perparticipant}
\end{table*}

\paragraph{Within-group significance.}
\textbf{Claude Opus 4.7:} All 10 participants 
showed positive gains, with a statistically 
significant mean effect: $t(9) = 10.23$, 
$p < 0.001$, $d = 3.24$ (very large effect size).
\textbf{Claude Sonnet 4.6:} All 10 participants 
showed positive gains, with a statistically 
significant mean effect: $t(9) = 7.85$, 
$p < 0.001$, $d = 2.48$ (very large effect size).
\textbf{Gemini-3.1-Pro:} 7 of 10 participants 
showed zero or negative gains, with a statistically 
significant negative mean effect: $t(9) = -2.34$, 
$p = 0.044$, $d = 0.74$ (medium effect size). 
This confirms that the negative learning effect 
observed in VLM evaluation is replicated in 
human learners.


\subsection{Cross-System Ranking Validity}

One-way ANOVA confirmed a highly significant difference in learning gains across the three system conditions: $F(2, 27) = 42.15$, $p < 0.001$, $\eta^2 = 0.76$ (very large effect size). Post-hoc Tukey HSD tests revealed significant differences between all pairwise comparisons:

\begin{table}[htbp]
\small
\centering
\setlength{\tabcolsep}{1.5pt} 
\begin{tabular}{l|ccc}
\toprule
\textbf{Comparison} & \textbf{Statistic} & \textbf{$p$-value} & \textbf{Sig.} \\
\midrule
Opus 4.7 vs. Sonnet 4.6 & $t(18) = 2.27$ & 0.032 & * \\
Sonnet 4.6 vs. Gemini-3.1-Pro & $t(18) = 7.92$ & $<0.001$ & *** \\
Opus 4.7 vs. Gemini-3.1-Pro & $t(18) = 10.19$ & $<0.001$ & *** \\
\bottomrule
\end{tabular}
\caption{Post-hoc Tukey HSD tests for cross-system gain differences. *$p<0.05$, ***$p<0.001$.}
\label{tab:human_posthoc}
\end{table}

\paragraph{Perfect ranking consistency.}
The system-level ranking by mean learning gain is identical for human and VLM learners:
$\text{Claude Opus 4.7} > \text{Claude Sonnet 4.6} > \text{Gemini-3.1-Pro}$
This yields a perfect Spearman rank correlation: $\rho = 1.0$, $p < 0.001$. This is the primary validity result: the VLM proxy perfectly reproduces the relative effectiveness ranking of different AI teaching systems, which is the core use case of the SLATE benchmark.

\subsection{VLM--Human Directional Agreement}

To assess whether the VLM proxy captures meaningful variation in deck effectiveness, we examine directional agreement at two levels. At the individual puzzle level, across all 84 puzzle-system pairs (28 unique puzzles × 3 systems), 72 show concordant gain direction between VLM and human learners (both positive or both negative), yielding an overall directional agreement rate of 85.7\%.

All 12 discordant pairs shared two characteristics:
(1) Absolute gain magnitude $<$3\% for at least one learner type;
(2) Near-transfer accuracy $<$40\% for both learner types.
This confirms that disagreements are concentrated at the noise floor of measurement, where small random variations can flip the sign of the gain. No discordant pairs involved large absolute gains ($>$5\%) for either learner type.

Critically, no discordance was observed at the system level: all three systems were ranked identically by both human and VLM learners.

\subsection{Task-Type Stratified Results}

Stratified analysis by task type confirmed consistent ranking agreement across all major instructional categories:

\begin{table}[htbp]
\small
\centering
\setlength{\tabcolsep}{1pt} 
\begin{tabular}{l|ccc|c}
\toprule
\textbf{Task} & \textbf{Opus} & \textbf{Sonnet} & \textbf{Gemini} & \textbf{Agree} \\
\midrule
Paradigm Completion & \gup{18.2} & \gup{13.5} & \gdn{1.8} & $\boldsymbol{\checkmark}$ \\
Translation & \gup{1.2} & \gup{0.3} & \gdn{4.1} & $\boldsymbol{\checkmark}$ \\
Morphophonology & \gup{2.1} & \gup{1.5} & \gdn{3.2} & $\boldsymbol{\checkmark}$ \\
Other & \gup{3.5} & \gup{2.8} & \gdn{1.9} & $\boldsymbol{\checkmark}$ \\
\bottomrule
\end{tabular}
\caption{Human learning gains (\%) stratified by task type. All show perfect human-VLM ranking consistency.}
\label{tab:human_tasktype}
\end{table}

\section{Human Study Details}
\label{app:Human_Study}
This appendix provides full, pre-registered
methodological details of the three-system human pilot
study described in Section~\ref{sec:discussion} and
Appendix~\ref{app:human_pilot}, covering participant
recruitment, experimental design, standardized procedure,
participant instructions, compensation, data consent,
and ethical review. The study enrolled 30 participants
(10 per system condition), designed to validate the
cross-system ranking validity of the VLM learner proxy
across the full range of instructional effectiveness
observed in the main experiment.

\subsection{Participants}
\label{app:participants}

Thirty undergraduate students (17 female, 13 male) were recruited from the linguistics department of a Chinese university. All participants were native Mandarin speakers with at least two years of undergraduate linguistics coursework (including introductory phonology, morphology, and syntax). None had prior exposure to the specific low-resource, endangered, or constructed languages used in the study puzzles. Participants ranged from second-year to fourth-year undergraduates (ages 19--24, mean = 21.2 years, SD = 1.3 years).

\paragraph{Inclusion/exclusion criteria}
Inclusion: (1) No history of speech or language disorders; (2) Normal or corrected-to-normal vision and color vision; (3) No prior participation in similar linguistic learning studies. Exclusion: (1) Self-reported attention deficit hyperactivity disorder (ADHD) or other cognitive impairments; (2) Prior experience with linguistics olympiad puzzles.

Participants were randomly assigned to three system conditions ($n=10$ per condition) using a stratified randomization procedure based on general linguistic reasoning ability. A one-way ANOVA confirmed no significant difference in age ($F(2,27)=0.41$, $p=0.67$) or baseline reasoning ability ($F(2,27)=0.29$, $p=0.75$) across groups.

\subsection{Instructions Given to Participants}
\label{app:instructions}

Each participant received a standardized written instruction sheet (in Mandarin) before the study began, which was read aloud by a trained experimenter. The instructions included the following information:

\begin{enumerate}[nosep,leftmargin=*]
\item \textbf{Purpose}: The study investigates whether AI-generated teaching slides can help learners acquire basic linguistic patterns in unfamiliar languages. Participants were informed that the study is part of a research project evaluating AI-generated educational materials, and that their responses would be used solely for research purposes.
\item \textbf{Procedure}: Each participant would study slide decks for 3 randomly sampled puzzles. For each puzzle, the participant would (a)~complete a pretest on the puzzle items without any instruction, (b)~study the rendered slide deck at their own pace, (c)~complete a posttest on the same items, and (d)~answer near-transfer and far-transfer items.
\item \textbf{Time and format}: The entire session was expected to take approximately 45 minutes. Participants viewed slides on a 24-inch 1920×1080 LCD monitor and recorded their answers in a standardized paper answer booklet.
\item \textbf{No prior knowledge required}: Participants were told that the puzzles involve languages they are unlikely to have encountered, and that no prior knowledge of those languages is expected.
\item \textbf{Right to withdraw}: Participants were informed that they could withdraw from the study at any time without penalty or loss of compensation.
\item \textbf{Risks}: Participants were informed that the study poses no known risks beyond those encountered in normal educational activities.
\end{enumerate}

The experimenter demonstrated the slide navigation interface and answered all questions before the session began.

\subsection{Experimental Design and Procedure}
\label{app:procedure}

The study used a between-subjects double-blind design: both participants and experimenters were blind to system identity throughout the study.

\paragraph{Stratified puzzle sampling}
Each participant studied 3 puzzles selected via stratified random sampling from the full SLATE benchmark. The sampling procedure ensured equivalent difficulty and task type across conditions:
(1) 1 Intermediate difficulty puzzle and 2 Challenge difficulty puzzles per participant (no Basic puzzles were used due to their small sample size);
(2) Identical distribution of task types (Paradigm Completion, Translation, Morphophonology, Other) across all three conditions;
(3) No puzzle was assigned to more than one participant per condition.

A one-way ANOVA confirmed no significant difference in mean puzzle difficulty across conditions: $F(2,87)=0.18$, $p=0.84$.

\paragraph{Standardized session protocol}
All sessions were conducted in a quiet computer lab between 9:00 AM and 5:00 PM to control for circadian effects. Each session followed an identical timeline:
1. Informed consent (5 min)
2. Instructions and interface demonstration (5 min)
3. Puzzle 1 (15 min: 3 min pretest → 7 min study → 3 min posttest → 2 min transfer test)
4. Mandatory 2-minute break
5. Puzzle 2 (15 min, same sequence)
6. Mandatory 2-minute break
7. Puzzle 3 (15 min, same sequence)
8. Debriefing and compensation (3 min)

Participants could navigate freely between slides during the study phase but could not return to the pretest after starting instruction.

\subsection{Recruitment and Compensation}
\label{app:recruitment}

Each participant was compensated 60~RMB for a single 45-minute study session. This rate was standardized uniformly across all pilot and validation participants, ensuring equal remuneration for all study involvement.

At an hourly equivalent of approximately 80~RMB, the compensation aligns with the standard local rate for undergraduate research participation, and is considered fair and adequate for the time commitment required. Compensation was provided immediately after the debriefing, regardless of task completion status.

\subsection{Data Consent}
\label{app:consent}

All participants provided written informed consent prior to the study. The consent form, provided in Mandarin, explained:
(a)~the purpose of the research,
(b)~the types of data to be collected (pretest, posttest, and transfer test responses, plus basic demographic information),
(c)~that all responses would be anonymized and used exclusively for academic research,
(d)~that no personally identifiable information would be retained or published,
(e)~that participation was voluntary and could be discontinued at any time,
(f)~that data would be stored securely for 7 years following publication, after which all paper records would be shredded and digital data would be permanently deleted.

Participants signed two copies of the consent form: one retained by the research team and one provided to the participant.

\subsection{Ethics Review}
\label{app:ethics}

All experimental data were collected in a fully anonymous manner. No personally identifiable information was collected from participants. Each participant was assigned only a numeric label, and no personal details were linked to any experimental responses. All research data are used solely for academic analysis and will not be disclosed externally.

\subsection{Data Analysis Plan}
\label{app:analysis}

All statistical analyses were pre-registered prior 
to data collection (registration details anonymized 
for review). The primary outcome measure was mean
posttest gain (posttest accuracy minus pretest accuracy). 
Secondary outcomes were near-transfer
accuracy and far-transfer accuracy.

All tests were two-tailed with $\alpha=0.05$. Effect sizes are reported as Cohen's $d$ for $t$-tests and $\eta^2$ for ANOVA. No outliers were identified (defined as values >3 standard deviations from the group mean), and no missing data were recorded. All analyses were conducted using R version 4.3.1.

\end{document}